\documentclass[11pt]{article}
\usepackage[preprint]{neurips_2024}
\makeatletter
\renewcommand{\@noticestring}{}
\makeatother
\usepackage[utf8]{inputenc}
\usepackage{amsmath,amsfonts}
\usepackage{amssymb}
\usepackage{algorithmic}
\usepackage{algorithm}
\usepackage{array}
\usepackage{subfig}
\usepackage{textcomp}
\usepackage{url}
\usepackage{hyperref}
\usepackage{cleveref}
\providecommand{\doi}[1]{\href{https://doi.org/#1}{\nolinkurl{doi:#1}}}
\usepackage{verbatim}
\usepackage{graphicx}
\usepackage{booktabs}
\usepackage{cancel}
\usepackage{enumitem}
\usepackage[most]{tcolorbox}
\usepackage{pgfplots}
\usepgfplotslibrary{groupplots}
\pgfplotsset{compat=1.18}

\usepackage{tikz}
\usetikzlibrary{arrows.meta, positioning, fit, calc, backgrounds, shapes.geometric}

\usepackage[]{todonotes}

\newtcolorbox[auto counter, Crefname={Example}{Examples}]{examplebox}[2][]{%
  enhanced,
  breakable,
  colback=blue!3!white,
  colframe=blue!45!black,
  boxrule=0.5pt,
  borderline west={2pt}{0pt}{blue!55!black},
  arc=1pt,
  left=8pt,
  right=8pt,
  top=6pt,
  bottom=6pt,
  fonttitle=\bfseries,
  coltitle=black,
  colbacktitle=blue!10!white,
  attach boxed title to top left={xshift=6pt,yshift=-2mm},
  boxed title style={
    boxrule=0pt,
    sharp corners,
    colback=blue!10!white
  },
  title={Example~\thetcbcounter: #2},
  #1
}

\begin{document}
%
\title{Active Inference for Physical AI Agents\\[0.5em]
{\Large\textnormal{An Engineering Perspective}}}

\author{
  Bert de Vries \\
  Department of Electrical Engineering \\
  Eindhoven University of Technology \\
  the Netherlands \\
  \texttt{bert.de.vries@tue.nl} \\[0.5em]
  \normalfont\small March 21, 2026 \,---\, Version 1.0
}

\maketitle

\begin{abstract}
Physical AI agents, such as robots and other embodied systems operating under tight and fluctuating resource constraints, remain far less capable than biological agents in open-ended real-world environments. This paper argues that Active Inference (AIF), grounded in the Free Energy Principle, offers a principled foundation for closing that gap. We develop this argument from first principles, following a chain from probability theory through Bayesian machine learning and variational inference to active inference and reactive message passing. From the FEP perspective, systems that maintain their structural and functional integrity over time can, under suitable assumptions, be described as if they minimize variational free energy (VFE), and AIF operationalizes that principle by unifying perception, learning, planning, and control within a single computational objective rather than separate engineered subsystems. We show that VFE minimization is naturally realized by reactive message passing on a factor graph, where inference emerges from local, parallel computations. This realization is not merely convenient: it is well matched to the defining constraints of physical operation, including hard temporal deadlines, asynchronous data arrival, fluctuating power budgets, and changing environmental composition. Because reactive message passing is event-driven, interruptible, and locally adaptable, performance can degrade gracefully under reduced resources while the model structure can adjust online as relevant entities and relations change. We further show that, under suitable coupling and coarse-graining conditions, coupled AIF agents can be described as higher-level AIF agents, yielding a computationally homogeneous architecture that uses the same message-passing primitive across scales. We do not present benchmark comparisons with existing methods; our contribution is to make the theoretical and architectural case accessible to the engineering community.
\end{abstract}

\tableofcontents
%
%
%
%
%

\section{Introduction}\label{sec:introduction}

RoboCup\footnote{\url{https://www.robocup.org/}} is an international research and education initiative that uses robot competitions as a benchmark problem to advance artificial intelligence, robotics, and autonomous multi-agent systems. The ultimate goal of RoboCup is stated as follows:\footnote{\url{https://www.robocup.org/objective}}
``\emph{By the middle of the 21st century, a team of fully autonomous humanoid robot soccer players shall win a soccer game, complying with the official rules of FIFA, against the winner of the most recent World Cup.}'' In October 2025, the RoboCup 2025 Finals for the category ``Humanoid Robot Play Football''\footnote{\url{https://humanoid.robocup.org/}} featured a match between two teams using Booster T1 robots.\footnote{\url{https://www.booster.tech/booster-t1}} An impression of that match can be viewed at YouTube.\footnote{\url{https://youtu.be/3Gyx-zT4gog?si=2MKeUGpRC2s-7qFD}}

First, let us commend and pay due respect to the skill, ingenuity, and sustained effort of the human engineering teams behind today’s robot football systems. Their achievements are substantial and hard-won. At the same time, intellectual honesty is required. A football team of human toddlers would likely defeat the current world champion humanoid robot team. The capability gap between contemporary large language model (LLM)–based AI systems, which rival or exceed human expert performance in tasks such as document processing and code development, and physical AI systems such as autonomous football robots is striking.

Let us consider the technology underlying contemporary robot football teams. Their development draws on state-of-the-art methods from disciplines such as control theory, signal processing, robotics, machine learning, and communications, representing decades of theoretical advances and large-scale engineering effort.

Now contrast the football skill levels of these robotic systems with those of an elite human football player, such as Kylian Mbapp\'e.\footnote{\url{https://youtu.be/BgL5bKxDOsE?si=nXYVAH2qZ42IMlia}} Mbapp\'e does not explicitly employ any knowledge of control theory or reinforcement learning. Instead, his brain and body operate under the laws of physics in a way that, through prolonged interaction with the environment, gives rise to football skills that vastly exceed those of current humanoid robots.

In fact, the magnitude of this performance gap is so large that it raises the question of whether continued development of control and learning algorithms is the most promising path toward humanoid robot teams capable of defeating the human world champions by 2050. Instead, it may be more fruitful to study how physical processes in biological brains give rise to information processing mechanisms that, through environmental interactions, produce exceptional sensorimotor skills.

We used the robot football example solely to illustrate the huge performance gap between human skill levels and current physical AI agents that must operate under the constraints of real-world embodiment. In this paper, our broader interest is to introduce an alternative approach to developing physical, agentic AI systems whose performance is competitive with that of humans.

\begin{figure}[t]
    \centering
    \includegraphics[width=\columnwidth]{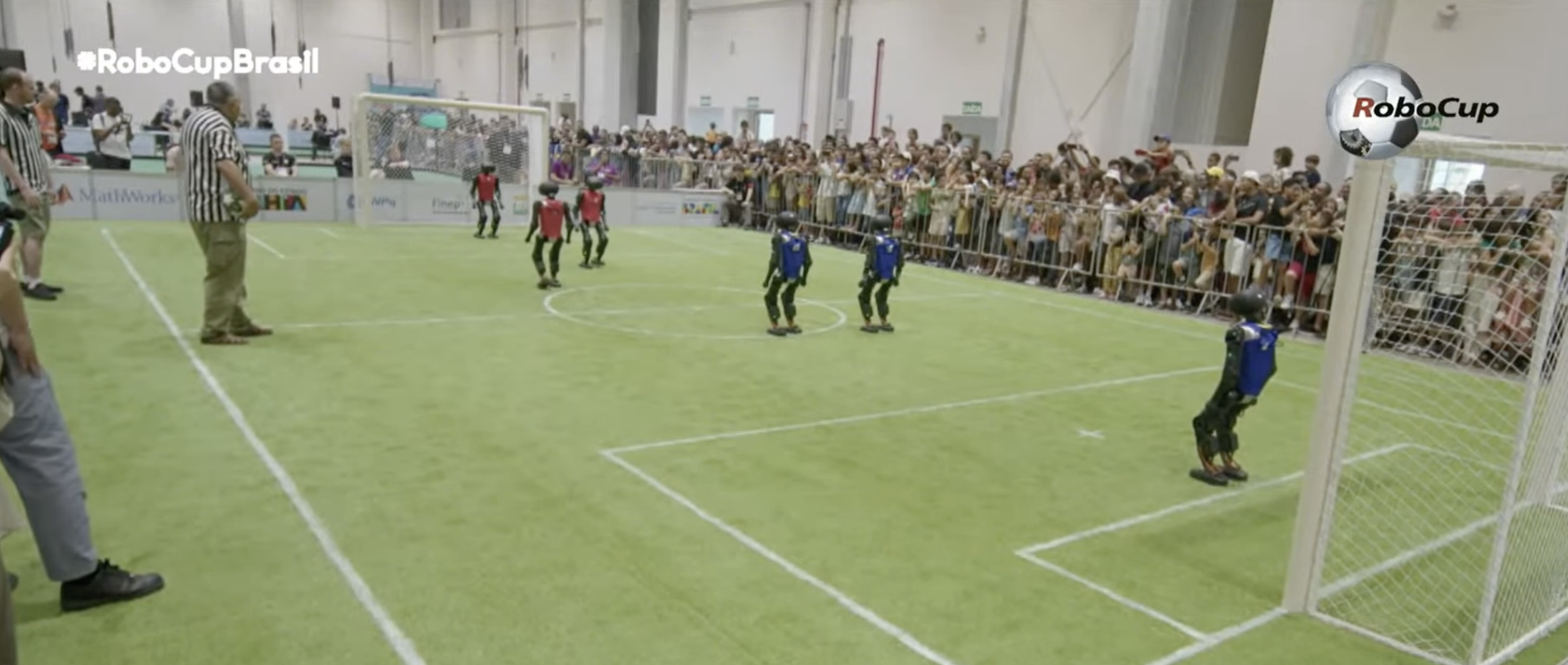}
    \caption{Screenshot from the RoboCup 2025 Finals in Brazil (``Humanoid Robot Play Football'' category), in which China's Booster Robotics team won the tournament. The robots played fully autonomously. Video: ``Humanoid Robot Play Football at RoboCup 2025 Finals in Brazil,'' uploaded by Chris Wabs on 20 July 2025. Available at \url{https://youtu.be/3Gyx-zT4gog?si=1kwZT1ikCm5pl2Lq}.}
    \label{fig:robocup-2025}
\end{figure}

Around twenty years ago, a unifying framework, known as the Free Energy Principle (FEP), was introduced for describing information processing in brains as a physical process \citep{fristontheory2005,fristonfreeenergy2009}. Since its introduction, the FEP has been further developed into a general least-action principle for the self-organizing dynamics of natural systems that maintain their structural and functional integrity \citep{fristonfree2022}. A comprehensive overview of the current state of the theory is provided in \citep{fristonfree2019,fristonpath2023}.

A defining feature of the FEP is its interpretation of the brain as a probabilistic generative model of sensory observations, in which all information processing is cast as the minimization of Variational Free Energy (VFE). Within this framework, cognitive processes such as perception, control, planning, learning, decision-making, attention, habit formation, exploration, and imagination are no longer treated as separate cognitive faculties, but arise as consequences of a single inferential principle. From an engineering perspective, this unification is conceptually powerful and potentially highly attractive, as these functions are typically addressed by fundamentally different methodologies across control, signal processing, machine learning, and artificial intelligence.

To distinguish the actual physical process from the overarching theoretical framework, the VFE minimization process in the brain is commonly referred to as Active Inference (AIF), although this terminology will be refined later in the paper. From an engineering perspective, AIF is of particular interest as a potential foundation for developing \emph{synthetic} physical AI systems that can autonomously acquire skills through interaction with their environment.

Despite its conceptual appeal, the literature on FEP and AIF remains difficult for engineers to penetrate. The aim of this paper is therefore to present AIF from an engineering viewpoint and to clarify why it constitutes a promising paradigm for researchers in fields such as robotics and other embodied AI systems. 

\begin{figure*}[t]
\centering
\resizebox{\textwidth}{!}{%
\begin{tikzpicture}[>=Stealth, every node/.style={font=\small}]

\tikzstyle{stagebox}=[draw, thick, rounded corners=6pt, minimum width=2.8cm,
    minimum height=1.6cm, align=center, fill=teal!8, text=teal!80!black,
    font=\normalsize\bfseries]

\node[stagebox] (PT)  at (0, 0)    {Probability\\Theory};
\node[stagebox] (BML) at (4.2, 0)  {Bayesian\\Machine\\Learning};
\node[stagebox] (VI)  at (8.4, 0)  {Variational\\Inference};
\node[stagebox] (AIF) at (12.6, 0) {Active\\Inference};
\node[stagebox] (RMP) at (16.8, 0) {FFG /\\Reactive MP};

\draw[->, line width=1.5pt, black!60] (PT.east)  -- (BML.west);
\draw[->, line width=1.5pt, black!60] (BML.east) -- (VI.west);
\draw[->, line width=1.5pt, black!60] (VI.east)  -- (AIF.west);
\draw[->, line width=1.5pt, black!60] (AIF.east) -- (RMP.west);

\tikzstyle{annot}=[anchor=north, align=center, font=\footnotesize,
    text=black!70, text width=3.2cm]

\node[annot] at (PT.south)  [yshift=-0.3cm]
    {Consistent reasoning\\under uncertainty\\(sum and product rules)};
\node[annot] at (BML.south) [yshift=-0.3cm]
    {Model-based learning\\and prediction\\(generative models)};
\node[annot] at (VI.south)  [yshift=-0.3cm]
    {Tractable approximate\\inference\\(VFE minimization)};
\node[annot] at (AIF.south) [yshift=-0.3cm]
    {Unified perception,\\learning, planning,\\and control};
\node[annot] at (RMP.south) [yshift=-0.3cm]
    {Distributed, anytime,\\fault-tolerant,\\resource-adaptive};

\tikzstyle{secref}=[anchor=south, align=center, font=\scriptsize,
    text=gray!70]

\node[secref] at (PT.north)  [yshift=0.2cm] {Section~\ref{sec:probability-theory}};
\node[secref] at (BML.north) [yshift=0.2cm] {Section~\ref{sec:bayesian-machine-learning}};
\node[secref] at (VI.north)  [yshift=0.2cm] {Section~\ref{sec:variational-inference}};
\node[secref] at (AIF.north) [yshift=0.2cm] {Section~\ref{sec:FEP-and-AIF-agents}};
\node[secref] at (RMP.north) [yshift=0.2cm] {Sections~\ref{sec:factor-graphs},~\ref{sec:physical-ai-as-aif}};

\end{tikzpicture}%
}
\caption{Conceptual pathway developed in this paper. Each stage builds on the previous one, culminating in a distributed, resource-adaptive architecture for physical AI agents. The annotations below each stage summarize the key property contributed at that level.}
\label{fig:pathway}
\end{figure*}
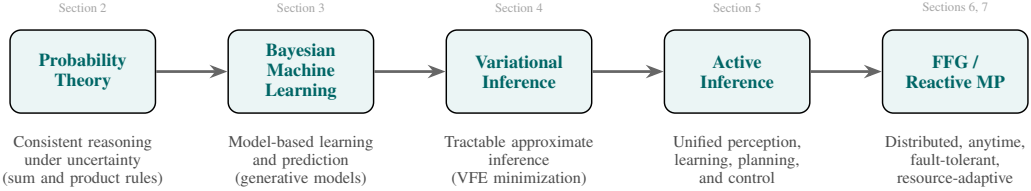

The paper develops its argument along the following chain (Fig.~\ref{fig:pathway}), where each step builds on the previous one:\footnote{Abbreviations used throughout: Active Inference (AIF), Bayesian Machine Learning (BML), Constrained Bethe Free Energy (CBFE), Expected Free Energy (EFE), Free Energy Principle (FEP), Kullback-Leibler (KL), Probability Theory (PT), Reactive Message Passing (RMP), Variational Free Energy (VFE), Variational Inference (VI).}
\begin{enumerate}[leftmargin=2em]
    \item \textbf{Probability Theory (PT).} We adopt a Bayesian interpretation of probabilities as degrees of belief and review the axiomatic derivation of the sum and product rules \citep{coxprobability1946,jaynesprobability2003}.
    \item \textbf{Bayesian Machine Learning (BML).} A full commitment to PT for learning from data. Bayes rule is the fundamental learning mechanism; model performance is scored by Bayesian model evidence. BML is principled but often computationally intractable.
    \item \textbf{Variational Inference (VI).} VFE minimization as a computationally tractable alternative to exact Bayesian inference, with deep roots in statistical physics \citep{feynmanslow1955,lanczosvariational1986} and axiomatic foundations in the maximum entropy principle.
    \item \textbf{Active Inference (AIF).} A full commitment to VFE minimization as the sole ongoing process in physical agents that interact with their environment through both actions and sensations.
    \item \textbf{Factor Graphs and Reactive Message Passing (RMP).} VFE minimization realized as distributed, event-driven message passing on a factor graph, naturally equipped to operate under fluctuating data, time, and power resources.
\end{enumerate}
In short, this paper aims to provide a clear pathway toward understanding and appreciating active inference as an engineering technology for the development of physical, agentic AI systems. We hope that presenting active inference from this perspective will encourage greater interest and adoption within the engineering community. The early sections on PT, BML and VI are deliberately paced to be accessible to a broad readership. Readers already familiar with VI may skip ahead to Section~\ref{sec:FEP-and-AIF-agents}, where the development of active inference begins.

\section{Probability Theory}\label{sec:probability-theory}

We begin with a review of probability theory (PT) to establish notation and to highlight the specific interpretation, Bayesian probability, on which the remainder of the paper relies. Due to space limitations, we focus on conceptual understanding rather than formal completeness. 

A classic way to introduce probability is to ask a student in a classroom to guess the number $x$ on the back of the blackboard, assuming $x$ is an integer between 1 and 100. Different students may give different answers because the uncertainty of their state of knowledge about $x$ can be expressed as
\begin{equation}\label{eq:student-belief}
    p(x=a) = \begin{cases}
        1/100 & \text{if } a \in \{1,2,\ldots,100\}  \\
        0 & \text{otherwise}
    \end{cases}\,,
\end{equation}
where $0\leq p(A) \leq 1$ represents the degree of belief about the truth of statement (or event) $A$. If I wrote the number $57$ on the blackboard, \emph{my} state of knowledge about $x$ can be expressed as
\begin{equation}\label{eq:bert-belief}
    p(x=a) = \begin{cases}
        1 & \text{if } a = 57 \\
        0 & \text{otherwise}
    \end{cases}\,.
\end{equation}
After turning the blackboard and revealing the number to the classroom, the students' state of knowledge changes from \eqref{eq:student-belief} to \eqref{eq:bert-belief}. 

Apparently, probability distributions provide a convenient way to represent a state of knowledge, in particular a degree of belief in an event (here: $x=a$) that can be either true or false. This interpretation of probability as a degree of belief is commonly referred to as the Bayesian interpretation. Importantly, the degree of belief about an event may change even though the underlying physical number on the blackboard ($x=57$) itself remains unchanged. In the present example, updating the state of knowledge after the reveal was trivial; however, in most practical situations, it is far less clear how the revelation of (partial) information should update one’s beliefs.

In a groundbreaking 1946 paper, the correct calculus for ``rational'' updating degrees of belief (probabilities) when new information becomes available was derived \citep{coxprobability1946}. In developing this calculus for rational information processing, only some very agreeable assumptions (``axioms'') were made, including:
\begin{itemize}
    \item[\textbf{C1}] \textit{Real-valued beliefs.}
    Degrees of belief are represented by real numbers between $0$ and $1$.

    \item[\textbf{C2}] \textit{Consistency.}
    Plausibility assessments are consistent: if $A$ becomes more plausible under new
    information $B$, the assigned degree of belief increases accordingly; and if belief
    in $A$ exceeds belief in $B$ which exceeds belief in $C$, then belief in $A$ must
    exceed belief in $C$.

    \item[\textbf{C3}] \textit{Logical closure.}
    Logical equivalences are preserved: if the belief in an event can be inferred in two
    different ways, e.g.\ by updating on $I_1$ then $I_2$ or vice versa, the two
    paths must agree on the resulting belief.
\end{itemize}

Cox derived that if 1--3 holds, then the only correct way to update probabilities must proceed via the sum and product rules:
\begin{subequations}\label{eq:sum-and-product-rules}
 \begin{align}
    p(A+B|I) &= p(A|I) + p(B|I) - p(A,B|I) \quad &&\text{(sum rule)} \label{eq:sum-rule} \\
    p(A,B|I) &= p(A|B,I) p(B|I) &&\text{(product rule)} \label{eq:product-rule}
\end{align}   
\end{subequations}
In \eqref{eq:sum-and-product-rules}, $p(A | I)$ denotes the conditional probability of $A$ given that $I$ is true. In $p(A + B)$, the plus-sign should be interpreted as the logical-OR between events $A$ and $B$, while $p(A,B)$ denotes the joint probability, corresponding to the logical-AND of $A$ and $B$.

The important conclusion is that any attempt to manipulate degrees of belief using a calculus that is inconsistent with \eqref{eq:sum-and-product-rules} risks violating at least one of Cox’s three axioms. In other words, if we accept that Cox’s axioms should hold, then we \emph{must} process new information using the sum and product rules in \eqref{eq:sum-and-product-rules}. Therefore, all rational information processing calculations can be reduced to successive applications of these two rules alone. In the following sections, we will show that Bayes rule, machine learning, variational inference, and ultimately active inference all derive from these two rules.

Two useful additional rules can be derived from the sum and product rules. The first is Bayes rule, which is simply a rewrite of the product rule combined with a symmetry argument. Since the conjunctions $A \land B$ and $B \land A$ evaluate to the same truth value, it follows that $p(A,B) = p(B,A)$, and consequently
\begin{equation}\label{eq:bayes-rule}
p(B | A) = \frac{p(A | B) p(B)}{p(A)} \quad \text{(Bayes rule)}\,.
\end{equation}
The significance of this relationship for machine learning will be discussed in the next section. 

Secondly, the Law of Total Probability (LTP) is an application of the sum rule to computing the probability of an event $A$ by summing over a set of related ``mutually exclusive and exhaustive'' events $B_i$. In particular, if $\{B_1,B_2,\ldots,B_n\}$ partitions the sampling space, then
\begin{equation}\label{eq:LTP}
 p(A) = \sum_{i=1}^n p(A,B_i) \quad \text{(law of total probability)}\,.  
\end{equation}
The application of the LTP is commonly referred to as marginalization, and the resulting distribution $p(A)$ is called the marginal probability. The sum and product rules, together with Bayes rule and the LTP, form the core workhorses of all rational information processing.

Example~\ref{ex:disease-diagnosis} provides an instructive example that highlights both the power of PT and the pitfalls of relying on intuition instead of the sum and product rules.

\begin{examplebox}[label=ex:disease-diagnosis]{Disease Diagnosis}
A disease $D\in \{0,1\}$ has prevalence $p(D=1) = 0.01$. A diagnostic test $T\in \{0,1\}$ has sensitivity $p(T=1|D=1) = 0.95$ and specificity $p(T=0|D=0) = 0.85$. What is the probability that a patient who tests positive actually has the disease, i.e.\ $p(D=1|T=1)$?

Applying Bayes rule and the law of total probability yields
\begin{align}
    p(D=1|T=1) &= \frac{p(T=1|D=1)\,p(D=1)}{p(T=1)} \tag{Bayes} \\
    &= \frac{p(T=1|D=1)\,p(D=1)}{p(T=1|D=1)\,p(D=1)+p(T=1|D=0)\,p(D=0)} \tag{LTP} \\
    &= \frac{0.95 \times 0.01}{0.95 \times 0.01 + 0.15 \times 0.99}
     \approx 0.06 \,. \notag
\end{align}
Despite the high sensitivity, a positive result carries only a $6\%$ chance of disease. This unintuitive result is called the \emph{base-rate fallacy}: $p(D=1|T=1) \approx 0.06$ differs drastically from $p(T=1|D=1) = 0.95$. The low prevalence dominates the posterior. Conflating $p(A|B)$ with $p(B|A)$ is a common error with serious consequences in medicine and law; systematic application of Bayes rule is the only safeguard.

\end{examplebox}


\section{Bayesian Machine Learning}\label{sec:bayesian-machine-learning}

Bayesian Machine Learning (BML) represents a full commitment to PT for the learning (from data) and application of models. In principle, BML is a sound idea because any alternative would imply a machine learning discipline that violates Cox’s axioms.

A key insight is that machine learning is generally not possible without introducing assumptions that go beyond the observed data alone. In a BML context, these assumptions are encoded by a \emph{model} $m$, which defines a joint probability distribution over a set of model parameters $\theta$ and an observed data set $D = \{y_1, y_2, \ldots, y_N\}$. This joint distribution is specified
by the product of a likelihood function and a prior distribution over the model parameters, i.e.,
\begin{equation}\label{eq:BML-model}
\underbrace{p(D,\theta|m)}_{\text{model}} =
    \underbrace{p(D|\theta,m)}_{\text{likelihood}}
    \underbrace{p(\theta|m)}_{\text{prior}} \, .
\end{equation}

Both the likelihood function and the prior warrant some clarification. The distribution $p(D | \theta, m)$ admits two complementary interpretations. First, it can be viewed as the probability that a model $m$, with fixed parameter values $\theta$, generates a data set $D$. In this interpretation, $D$ is the variable and $p(D | \theta, m)$ is read as a \emph{data-generating distribution}.

In a machine learning context, however, the data $D$ are typically regarded as a fixed
set of known observations, while the parameter values $\theta$ are unknown. Under this
interpretation, $p(D | \theta, m)$ is considered a function of $\theta$, which is referred to as the \emph{likelihood function}. From a modeling perspective,
the specification of an appropriate likelihood function is a central design choice and
reflects the assumptions made by the practitioner about the data-generating process.

In addition to the likelihood function, the practitioner must specify their state of
knowledge about the model parameters $\theta$ \emph{prior} to observing the data $D$. This knowledge is encoded by a prior distribution $p(\theta|m)$. The explicit
specification of a prior is a key aspect in which BML differs from alternative machine learning frameworks, where the required assumptions are typically introduced in an implicit or non-probabilistic manner, for example, through regularization terms.

Once a model \eqref{eq:BML-model} has been specified, and a new data set $D$ has been made available, all subsequent processing proceeds by PT. The actual learning task involves evaluating
\begin{subequations}\label{eq:BML-calculations}
 \begin{align}
    \overbrace{p(D|m)}^{\text{evidence}} &= \int \overbrace{p(D|\theta,m) p(\theta|m)}^{\text{known from }\eqref{eq:BML-model}} \mathrm{d}\theta \label{eq:compute-evidence} \\
    \underbrace{p(\theta|D,m)}_{\text{posterior}} &= \frac{ p(D|\theta,m) p(\theta|m)}{\underbrace{p(D|m)}_{\text{known from }\eqref{eq:compute-evidence}}} \label{eq:compute-posterior}\,.
\end{align}   
\end{subequations}
Hence, the model evidence $p(D|m)$ is first computed by marginalization over the model parameters $\theta$, and then the posterior distribution $p(\theta|D,m)$ for the parameters follows from Bayes rule. 

Both $p(D|m)$ and $p(\theta|D,m)$ are quantities of central interest. Bayes rule in \eqref{eq:compute-posterior} describes how the data set $D$ updates the beliefs about model parameters from a prior distribution $p(\theta|m)$ to a posterior distribution $p(\theta|D,m)$ by consistent rational reasoning. Therefore, \emph{Bayes rule
constitutes the fundamental rule underlying learning from data.} Deviating from \eqref{eq:compute-posterior} would potentially break the Cox axioms.

The evidence $p(D|m)$ is very valuable in its own right as it scores the performance of the model assumptions \eqref{eq:BML-model}. Note first that, for a given data set, $p(D|m)$ evaluates to a scalar value in the interval $(0,1]$. The negative logarithm of this quantity, commonly referred to as the \emph{surprisal}, admits a decomposition into a \emph{complexity} term minus an \emph{accuracy} term (see the proof in \eqref{eq:proof-surprise}),
\begin{equation}\label{eq:CA-decomp}
    \underbrace{-\log \overbrace{p(D|m)}^{\text{evidence}}}_{\text{surprisal}} =
    \underbrace{\int p(\theta|D,m)\log \frac{p(\theta|D,m)}{p(\theta|m)}\,\mathrm{d}\theta}_{\text{complexity}}
    -
    \underbrace{\int p(\theta|D,m)\log p(D|\theta,m)\,\mathrm{d}\theta}_{\text{accuracy}} \, .
\end{equation}

The complexity term is a Kullback-Leibler (KL) divergence\footnote{\url{https://en.wikipedia.org/wiki/Kullback\%E2\%80\%93Leibler_divergence}}, which can be interpreted as a non-negative distance measure between the posterior and prior probability distributions. The complexity term reflects how much we had to change our beliefs from the prior to the posterior distribution by absorbing the data $D$ into our model. A good model should avoid unnecessary deviation from the prior, as belief updates correspond to forgetting prior information (and, in an active inference context, forgetting what sustained survival is undesirable). The accuracy term is the expected log-likelihood, where the expectation is taken with respect to the (posterior) parameter beliefs. A good model has high accuracy since we want to predict the data set $D$ well. Both minimizing complexity and maximizing accuracy are aligned with the maximization of the model
evidence $p(D|m)$. Taken together, models with high Bayesian evidence strike a favorable trade-off by achieving high accuracy (i.e., good data fit to new data $D$) while maintaining low complexity (i.e., do not forget past learned patterns). This complexity--accuracy trade-off will reappear as the central design criterion when we develop the variational free energy objective for active inference agents in Sections~\ref{sec:variational-inference} and~\ref{sec:control}.

While evidence $p(D|m)$ scores the performance of model $m$, it is the posterior distribution $p(\theta|D,m)$ that is usually needed in applications of a trained model. For instance, we can evaluate the state of knowledge about a future observation $y_\bullet$, given the data set $D$ and the model assumptions \eqref{eq:BML-model}, as
\begin{equation}\label{eq:BML-application}
    p(y_\bullet|D,m) = \int \underbrace{p(y_\bullet|\theta,m)}_{\text{from }\eqref{eq:BML-model}}\underbrace{p(\theta|D,m)}_{\text{from }\eqref{eq:compute-posterior}} \mathrm{d}\theta \,.
\end{equation}

Note that all information processing in BML, that is, learning via \eqref{eq:BML-calculations} and model application as in \eqref{eq:BML-application}, relies exclusively on PT. Thus, BML represents a commitment to
performing machine learning without violating the Cox axioms.

In an applied setting, a practitioner usually iterates over candidate model proposals by
evaluating the model evidence using \eqref{eq:compute-evidence} until a satisfactory
model is obtained. Once an acceptable model has been selected, the posterior
distribution over the model parameters is computed via \eqref{eq:compute-posterior}.
The resulting model can then be applied, for example as described in \eqref{eq:BML-application}. A full example of BML applied to predicting coin tosses is shown in Example~\ref{ex:coin-toss}.

If we accept the Cox axioms, then we should accept BML as our machine learning framework. Unfortunately, evaluating the evidence via \eqref{eq:compute-evidence} may be intractable. As an illustrative example, consider a (small) model with $80$ parameters, where each parameter can take $10$ possible values. Evaluating the evidence using \eqref{eq:compute-evidence} then requires a summation over $10^{80}$ terms, which is on the order of the number of electrons in the universe. If the evidence cannot be evaluated, then the posterior distribution cannot be computed via \eqref{eq:compute-posterior}, and consequently, the application step in \eqref{eq:BML-application} also becomes intractable. Therefore, while BML is formally the correct approach to machine learning, computational limitations have hindered its widespread adoption.

\begin{examplebox}[label=ex:coin-toss]{Coin Toss Prediction}

\textbf{Dataset.} We observe $D = \{1,0,1,1,0,0,1\}$ ($N=7$ tosses, $n=4$ heads). What is the probability that the next toss is heads?

\medskip
\textbf{Step 1 — Model design.} We design two candidate models that share the same Bernoulli likelihood,
\begin{equation*}
    p(D|\mu,m_k) = \mathrm{Binomial}(n | N, \mu) = \mu^n(1-\mu)^{N-n}\,,
\end{equation*}
but differ in their Beta prior over the bias parameter $\mu\in[0,1]$:
\begin{equation*}
    p(\mu|m_1) = \mathrm{Beta}(\mu|100,500)\,,\quad
    p(\mu|m_2) = \mathrm{Beta}(\mu|8,13)\,.
\end{equation*}
Model $m_1$ encodes a strong belief that the coin is biased toward tails (prior mean $\frac{100}{600}\approx 0.17$); $m_2$ is more diffuse with a prior mean $\frac{8}{21}\approx 0.38$.

\medskip
\textbf{Step 2 — Evidence and model selection} (see Appendix~\ref{app:coin-toss-proof} for proof)\textbf{.} The model evidence is
\begin{equation}\label{eq:coin-toss-evidence-1}
    p(D|m_k) = \frac{B(n+\alpha_k,\;N-n+\beta_k)}{B(\alpha_k,\beta_k)}\,,
\end{equation}
where $B(\cdot,\cdot)$ is the beta function. The evidence is a function of both model choices $(\alpha_k,\beta_k)$ and observations $(N,n)$. Evaluating numerically:
$p(D|m_1)\approx 4.6\times 10^{-4}$ and $p(D|m_2)\approx 4.8\times 10^{-3}$.
Model $m_2$ has roughly ten times higher evidence and is selected.

\smallskip
\begin{tikzpicture}
\begin{groupplot}[
    group style={group size=2 by 1, horizontal sep=1.4cm},
    width=0.47\linewidth, height=3.2cm,
    xlabel={$\mu$},
    ymin=0, ymax=1.18,
    ytick=\empty,
    axis lines=left,
    tick style={thin},
    tick label style={font=\small},
    xlabel style={font=\small},
    title style={font=\small\bfseries},
    legend style={font=\tiny, at={(0.97,0.97)}, anchor=north east,
                  row sep=-3pt, inner sep=2pt},
    legend cell align=left,
    every axis plot/.append style={line width=0.9pt},
]
\nextgroupplot[
    title={Model $m_1$: $\mathrm{Beta}(\mu|100,500)$},
    domain=0.05:0.40, samples=500, xmin=0.05, xmax=0.40,
    xtick={0.1,0.2,0.3,0.4},
]
\addplot[blue!70!black, dashed, thick]
    {exp(99*ln(x/0.16555) + 499*ln((1-x)/0.83445))};
\addlegendentry{prior}
\addplot[green!60!black, dotted, thick]
    {exp(4*ln(x/0.57143) + 3*ln((1-x)/0.42857))};
\addlegendentry{likelihood}
\addplot[red!70!black, solid, thick]
    {exp(103*ln(x/0.17025) + 502*ln((1-x)/0.82975))};
\addlegendentry{posterior}
\nextgroupplot[
    title={Model $m_2$: $\mathrm{Beta}(\mu|8,13)$},
    domain=0.05:0.95, samples=400, xmin=0.05, xmax=0.95,
    xtick={0.2,0.4,0.6,0.8},
]
\addplot[blue!70!black, dashed, thick]
    {exp(7*ln(x/0.36842) + 12*ln((1-x)/0.63158))};
\addlegendentry{prior}
\addplot[green!60!black, dotted, thick]
    {exp(4*ln(x/0.57143) + 3*ln((1-x)/0.42857))};
\addlegendentry{likelihood}
\addplot[red!70!black, solid, thick]
    {exp(11*ln(x/0.42308) + 15*ln((1-x)/0.57692))};
\addlegendentry{posterior}
\end{groupplot}
\end{tikzpicture}

\noindent{\scriptsize Note: the horizontal axis for model $m_1$ is zoomed to $[0.05, 0.40]$ because the strong prior concentrates all three distributions in a narrow range.}
\smallskip

\medskip
\textbf{Step 3 — Posterior for $m_2$} (see Appendix~\ref{app:coin-toss-proof}, \eqref{eq:coin-toss-derivation})\textbf{.} Because the Beta prior is conjugate to the Binomial likelihood, the posterior is Beta-distributed:
\begin{equation*}
    p(\mu|D,m_2) = \mathrm{Beta}(\mu | n+\alpha_2,\; N-n+\beta_2) = \mathrm{Beta}(\mu|12,\,16)\,.
\end{equation*}

\medskip
\textbf{Step 4 — Prediction.} Marginalizing over the posterior of $m_2$:
\begin{equation*}
    p(x_\bullet=1|D,m_2)
    = \int_0^1 \mu\; p(\mu|D,m_2)\,\mathrm{d}\mu
    = \frac{n+\alpha_2}{N+\alpha_2+\beta_2}
    = \frac{4+8}{7+8+13} = \frac{12}{28} \approx 0.43\,.
\end{equation*}
The prediction equals the posterior mean $0.43$, reflecting a balanced fusion of a prior belief with mean $0.38$ and the observed frequency $n/N\approx 0.57$.

\end{examplebox}

\section{Variational Inference}\label{sec:variational-inference}

As discussed, when computational resources are limited, evaluating Bayesian evidence and the resulting posterior distribution may be intractable. An axiomatic framework for inference under constraints was developed in \citep{shoreaxiomatic1980}, later finessed in \citep{skillingclassic1989,catichaentropy2021}, closely analogous in spirit to Cox’s axiomatic derivation of PT.

In the context of a BML task, \citet{shoreaxiomatic1980} introduced a ranking functional $S[q]$ over candidate posterior distributions $q(\theta)$\footnote{For notational brevity, we drop the dependence on model $m$ in this section.}, defined relative to a prior model $p(D,\theta)$ and a newly imposed set of constraints. These constraints represent newly acquired information, such as observations in a data set, but may also include modeling restrictions that limit the admissible family of candidate distributions. For example, one may restrict attention to Gaussian posteriors $q(\theta)$. More generally, a constraint is any condition that affects the updating of beliefs from the prior to the posterior. \citet{shoreaxiomatic1980} required that the ranking functional satisfy the following axioms:
\begin{itemize}
    \item[\textbf{S1}] \textit{Uniqueness.}
    The updating rule must yield a unique posterior.

    \item[\textbf{S2}] \textit{Coordinate invariance.}
    Inference must be invariant under reparameterization.

    \item[\textbf{S3}] \textit{System independence.}
    Independent systems updated separately or jointly must lead to consistent results.

    \item[\textbf{S4}] \textit{Subset independence.}
    Constraints imposed on one subset must not affect inferences about disjoint subsets.
\end{itemize}

Intuitively, these axioms require that the posterior be determined solely by the constraints imposed, without introducing any unwarranted information. For example, S4 requires that calibrating a robot's camera must not alter beliefs about its microphone parameters, since the calibration data carries no information about the microphone. If these axioms are satisfied, then \citet{shoreaxiomatic1980} showed that the relative entropy functional is the unique ranking criterion that satisfies this requirement. Specifically, for a given set of observations $D$, the preferred posterior $q(\theta)$ is uniquely determined as the distribution that satisfies the imposed constraints while maximizing the \emph{relative entropy}
\begin{equation}\label{eq:relative-entropy}
S[q] = - \int q(\theta) \log \frac{q(\theta)}{p(D,\theta)} 
\mathrm{d}\theta \,.
\end{equation}
This inference method is known as the \emph{Maximum Entropy Principle} (MEP). While relative entropy is a central concept in information theory, its negative counterpart is known in statistical physics as the \emph{variational free energy} (VFE):
\begin{subequations}\label{eq:VFE}
\begin{align}
F[q] &= \int q(\theta) \log \frac{q(\theta)}{p(D,\theta)}
\mathrm{d}\theta \label{eq:VFE-def}\\
&= \underbrace{-\log p(D)}_{\substack{\text{surprisal} \\ \text{(problem repr. cost)}}}
+ \underbrace{\int q(\theta) \log \frac{q(\theta)}{p(\theta|D)}
\mathrm{d}\theta}_{\substack{\text{inference bound} \\ \text{(solution cost)}}} \label{eq:VFE-SB} \\
&= \underbrace{\int q(\theta) \log \frac{q(\theta)}{p(\theta)}
\mathrm{d}\theta}_{\text{complexity}} 
- \underbrace{\int q(\theta) \log p(D|\theta)
\mathrm{d}\theta}_{\text{accuracy}} \label{eq:VFE-CA} \\
&= \underbrace{\int q(\theta) \log \frac{1}{p(D,\theta)}\mathrm{d}\theta}_{\text{energy}} - \underbrace{\int q(\theta) \log \frac{1}{q(\theta)}\mathrm{d}\theta}_{\text{entropy}}  \,. \label{eq:VFE-EE} 
\end{align}
\end{subequations}

Eqs.~\eqref{eq:VFE-SB} and \eqref{eq:VFE-CA} follow from \eqref{eq:VFE-def} by applying the product rules $p(D,\theta) = p(\theta|D)\,p(D)$ and $p(D,\theta) = p(D|\theta)\,p(\theta)$, respectively. Since the second term in \eqref{eq:VFE-SB} is a Kullback–Leibler (KL) divergence and therefore non-negative, the variational free energy $F[q]$ constitutes an upper bound on the surprisal $-\log p(D)$. In the machine learning literature, the negative free energy $S[q] = -F[q]$ is commonly referred to as the \emph{evidence lower bound} (ELBO).

We also note that $F[q]$ admits a complexity–accuracy decomposition analogous to the decomposition of Bayesian model evidence in \eqref{eq:CA-decomp}, with the important distinction that \eqref{eq:VFE-CA} defines a functional of the approximate posterior $q(\theta)$, thereby enabling an explicit trade-off between complexity and inaccuracy in the choice of $q$.

\emph{Variational Inference} (VI) refers to selecting the posterior $q(\theta)$ by \emph{minimization} of $F[q]$ rather than by marginalization and Bayes rule as in \eqref{eq:BML-calculations}. Note that, by \eqref{eq:VFE-SB}, it follows that global minimization of VFE leads to
\begin{equation}\label{eq:argmin-F}
    q^*(\theta) = \arg\min_q F[q]
\end{equation}
where
\begin{subequations}
\begin{align}
    q^*(\theta) &= p(\theta|D) \label{eq:q*}\\
    F[q^*] &= -\log p(D) \label{eq:F[q*]}\,.
\end{align}
\end{subequations}
In other words, global minimization of the VFE with respect to the variational posterior $q$ recovers both the (optimal) Bayesian posterior in \eqref{eq:q*} as well as Bayesian evidence by \eqref{eq:F[q*]}. 

In practical applications, however, the minimization is typically restricted to a tractable variational family $\mathcal{Q}$, so that
\begin{equation}\label{eq:q-hat-argmin}
\hat{q}(\theta) = \arg\min_{q \in \mathcal{Q}} F[q] \,,
\end{equation}
where
\begin{subequations}
\begin{align}
\hat{q}(\theta) &\approx p(\theta | D) \label{eq:q-hat} \\
F[\hat{q}] &\approx -\log p(D) \label{eq:F[q-hat]} \,.
\end{align}
\end{subequations}
This procedure yields an approximate, yet computationally tractable, Bayesian solution.

We have a remarkable result. Exact Bayesian updating via \eqref{eq:BML-calculations} is optimal in the sense that it conforms to the Cox axioms, but it is often computationally intractable due to the marginalization over $\theta$ required in \eqref{eq:compute-evidence}. Variational inference circumvents this marginalization by re-framing Bayesian learning as the optimization problem in \eqref{eq:q-hat-argmin}, which is typically far more tractable computationally.

Moreover, if one augments the inference problem with constraints beyond data constraints, then \citet{shoreaxiomatic1980,skillingclassic1989,catichaentropy2021} provide a strong axiomatic motivation for treating VFE minimization under constraints as a principled approach to inference. Hence, VFE minimization is not merely a convenient approximation technique, but a principled (and in practice unavoidable) framework for consistent reasoning under uncertainty in the real-time conditions faced by physical AI systems. This leads to a key conceptual insight: \emph{In this constrained-inference view, Bayes rule appears as a special case of VI in which the only constraints encode the observed data and no restriction is imposed on the admissible family of posteriors.} VI is therefore more general than Bayes rule, because it accommodates additional constraints (computational, structural, or distributional) that real-world agents inevitably face.

To emphasize this point, the surprisal–bound decomposition in \eqref{eq:VFE-SB} can be interpreted as a decomposition into problem representation costs and solution costs. Specifically, surprisal quantifies how well the model represents the environment, while any practical use of the model in a solution necessarily incurs inference costs. An important consequence is that a model with a relatively poor problem representation (high surprisal), but paired with an efficient inference process (low solution cost), may achieve a lower VFE than a model with high Bayesian evidence but an expensive or inaccurate inference procedure. This implies that the model with the highest Bayesian evidence is not necessarily the most useful \emph{in practice}, since Bayesian evidence evaluates only the quality of the problem representation and ignores the computational cost of inference. Accordingly, the common interpretation of VFE as merely an upper bound on surprisal is incomplete. Instead, VFE provides a more principled performance criterion, as it jointly evaluates model fidelity and the computational cost of the inference process, which is an essential consideration for physical AI systems operating under real-time and resource constraints. 

This broader interpretation is reinforced by the energy--entropy decomposition in \eqref{eq:VFE-EE}, which links VFE to the free-energy functionals of statistical physics. More generally, variational free-energy principles are closely related to Jaynes' maximum-entropy principle and, through statistical mechanics, to the second law of thermodynamics. Variational inference should therefore not be viewed merely as an approximation to exact Bayesian inference, but as a fundamental principle of information processing.

An example of this VI workflow for Bayesian logistic regression is given in Example~\ref{ex:vi-logistic-regression}.
\begin{examplebox}[label=ex:vi-logistic-regression]{Variational Inference for Logistic Regression}

\textbf{Task.} Given labeled data
$D=\{(x_n,y_n)\}_{n=1}^{N}$ with feature vectors $x_n \in \mathbb{R}^M$
and binary labels $y_n \in \{0,1\}$, infer a classifier that predicts
$p(y_\bullet=1|x_\bullet,D)$.

\smallskip
\textbf{Step 1 — Model.} In \emph{logistic regression} we assume a model
\begin{equation}\label{eq:vlr-joint}
\begin{aligned}
    p(w) &= \mathcal{N}(w|0,\alpha^{-1}I)\,, \\
    p(y_n|x_n,w) &= \sigma(w^\top x_n)^{y_n}\big(1-\sigma(w^\top x_n)\big)^{1-y_n}\,.
\end{aligned}
\end{equation}
where $\sigma(a)=1/(1+\exp(-a))$ is the logistic sigmoid.

\smallskip
\textbf{Step 2 — Why VI is needed.} The posterior $p(w|D) = p(D,w) / p(D)$ is not available in closed form because the evidence
\[
    p(D)=\int p(w)\prod_{n=1}^{N} p(y_n|x_n,w)\,\mathrm{d}w
\]
involves a product of sigmoid terms. VI replaces this
intractable integration by minimization of
\begin{equation}
    F[q] = \int q(w)\log\frac{q(w)}{p(D,w)}\,\mathrm{d}w\,.
    \label{eq:vlr-vfe}
\end{equation}

\smallskip
\textbf{Step 3 — Gaussian variational posterior.} We impose the following (form) constraint on the posterior: 
\begin{equation} \label{eq:vlr-gaussian-posterior}
    q(w)=\mathcal{N}(w|m,S)\,.
\end{equation}
Substituting \eqref{eq:vlr-joint} and \eqref{eq:vlr-gaussian-posterior} into \eqref{eq:vlr-vfe} gives
\begin{align}
    F(m,S)
    &= \mathrm{KL}\!\left[\mathcal{N}(w|m,S)\,\big\|\,\mathcal{N}(w|0,\alpha^{-1}I)\right] \notag\\
    &\quad - \sum_{n=1}^{N} \mathbb{E}_{q(w)}\!\left[
    y_n \log \sigma(w^\top x_n)
    +(1-y_n)\log\big(1-\sigma(w^\top x_n)\big)\right].
    \label{eq:vlr-vfe-ms}
\end{align}
The first term is analytic, but the expected log-sigmoid terms are not.
In practice, one minimizes \eqref{eq:vlr-vfe-ms} by 
stochastic gradient estimates or by message passing as detailed in Section~\ref{sec:factor-graphs}.

\smallskip
\textbf{Step 4 — Prediction.} After optimization, predictions for a new
feature vector $x_\bullet$ are obtained by marginalizing over the
variational posterior:
\begin{equation}
    p(y_\bullet=1|x_\bullet,D) = \int p(y_\bullet=1|x_\bullet,w)p(w|D)\mathrm{d}w
    \approx \int \sigma(w^\top x_\bullet)\,q(w)\,\mathrm{d}w\,.
    \label{eq:vlr-predictive}
\end{equation}
Thus, the classifier depends not only on the posterior mean $m$, but also
on the uncertainty encoded in $S$.

\smallskip
\centering
\begin{tikzpicture}[x=1cm,y=1cm]
\begin{scope}[shift={(0,0)}]
    \draw[->,thin] (0,0) -- (5.2,0) node[right] {$w_1$};
    \draw[->,thin] (0,0) -- (0,3.2) node[above] {$w_2$};
    \draw[gray!50] (0,0) rectangle (5,3);
    \node[font=\small\bfseries] at (2.5,3.35) {variational approximation in weight space};
    \draw[densely dashed,gray!80] (1.4,1.35) ellipse (1.0 and 1.0);
    \draw[thick,blue!70!black,rotate around={18:(3.05,1.65)}]
        (3.05,1.65) ellipse (1.25 and 0.65);
    \draw[thick,red!70!black]
        plot[smooth cycle,tension=0.9] coordinates
        {(1.95,1.45) (2.45,2.45) (3.45,2.35) (4.15,1.7) (3.55,0.95) (2.55,0.9)};
    \fill[black] (3.05,1.65) circle (1.1pt);
    \node[font=\scriptsize,gray!80] at (1.35,2.55) {prior $p(w)$};
    \node[font=\scriptsize,red!70!black] at (3.85,2.65) {posterior $p(w|D)$};
    \node[font=\scriptsize,blue!70!black] at (3.75,0.65) {$q(w)=\mathcal{N}(m,S)$};
    \node[font=\scriptsize] at (3.25,1.25) {$m$};
\end{scope}
\begin{scope}[shift={(6.5,0)}]
    \node[font=\small\bfseries] at (2.5,3.15) {Bayesian predictive distribution};
    \begin{scope}
    \clip (0,0) rectangle (5,2.8);
    \foreach \ix in {0,...,24} {
        \foreach \iy in {0,...,13} {
            \pgfmathsetmacro{\cx}{0.1 + \ix * 0.2}
            \pgfmathsetmacro{\cy}{0.1 + \iy * 0.2}
            \pgfmathsetmacro{\z}{1.8*(\cx - 0.65*\cy - 1.8)}
            \pgfmathsetmacro{\zc}{min(6, max(-6, \z))}
            \pgfmathsetmacro{\sig}{1 / (1 + exp(-\zc))}
            \pgfmathsetmacro{\rcomp}{\sig < 0.5 ? 2*\sig : 1}
            \pgfmathsetmacro{\gcomp}{\sig < 0.5 ? 2*\sig : 2*(1-\sig)}
            \pgfmathsetmacro{\bcomp}{\sig < 0.5 ? 1 : 2*(1-\sig)}
            \definecolor{cellcolor}{rgb}{\rcomp,\gcomp,\bcomp}
            \fill[cellcolor] (\ix*0.2, \iy*0.2) rectangle ++(0.2,0.2);
        }
    }
    \end{scope}
    \draw[thick, black] (1.8,0) -- (3.625,2.8);
    \foreach \px/\py in {0.6/2.3, 0.9/1.8, 1.1/2.6, 1.4/2.0, 0.7/1.5, 1.6/2.4, 1.0/2.1, 1.3/1.6} {
        \fill[blue!80!black] (\px,\py) circle (2.0pt);
        \draw[white, line width=0.3pt] (\px,\py) circle (2.0pt);
    }
    \foreach \px/\py in {3.0/0.7, 3.4/1.4, 3.8/0.4, 4.1/1.1, 4.4/0.8, 3.6/1.7, 4.2/1.4, 3.2/0.2} {
        \fill[red!80!black] (\px,\py) circle (2.0pt);
        \draw[white, line width=0.3pt] (\px,\py) circle (2.0pt);
    }
    \draw[->,thin] (0,0) -- (5.2,0) node[right] {$x_1$};
    \draw[->,thin] (0,0) -- (0,3.0) node[above] {$x_2$};
    \node[font=\scriptsize, blue!70!black] at (0.7,0.35) {$y{=}0$};
    \node[font=\scriptsize, red!70!black] at (4.3,2.45) {$y{=}1$};
    \foreach \ib in {0,...,27} {
        \pgfmathsetmacro{\frac}{\ib/27}
        \pgfmathsetmacro{\rc}{\frac < 0.5 ? 2*\frac : 1}
        \pgfmathsetmacro{\gc}{\frac < 0.5 ? 2*\frac : 2*(1-\frac)}
        \pgfmathsetmacro{\bc}{\frac < 0.5 ? 1 : 2*(1-\frac)}
        \definecolor{barcolor}{rgb}{\rc,\gc,\bc}
        \fill[barcolor] (5.5, \ib*0.1) rectangle ++(0.25, 0.1);
    }
    \draw[thin] (5.5, 0) rectangle (5.75, 2.8);
    \node[font=\scriptsize, anchor=west] at (5.85, 0) {$0$};
    \node[font=\scriptsize, anchor=west] at (5.85, 1.4) {$0.5$};
    \node[font=\scriptsize, anchor=west] at (5.85, 2.8) {$1$};
    \node[font=\scriptsize, anchor=south, rotate=90] at (6.6, 0.7) {$p(y{=}1|x,D)$};
\end{scope}
\end{tikzpicture}

\end{examplebox}

\section{The Free Energy Principle and Active Inference}\label{sec:FEP-and-AIF-agents}

We now turn to \emph{physical AI agents}, i.e., systems that are embodied in and act through a physical body in the real world. 
A physical AI agent, 
\begin{itemize}
    \item receives sensory input from its environment,
    \item performs inference and decision making based on these observations, and
    \item generates physical actions that influence the environment through its actuators.
\end{itemize}

A defining property of a physical AI agent is the existence of a \emph{closed perception--action loop}, in which actions alter the environment that subsequently generates new sensory input. This loop can be abstracted as
\begin{equation*}
    \cdots\; \text{sensors} \;\rightarrow\; \text{inference} \;\rightarrow\; \text{control} \;\rightarrow\; \text{actuators} \;\rightarrow\; \text{world} \;\rightarrow\; \text{sensors} \;\cdots.
\end{equation*}

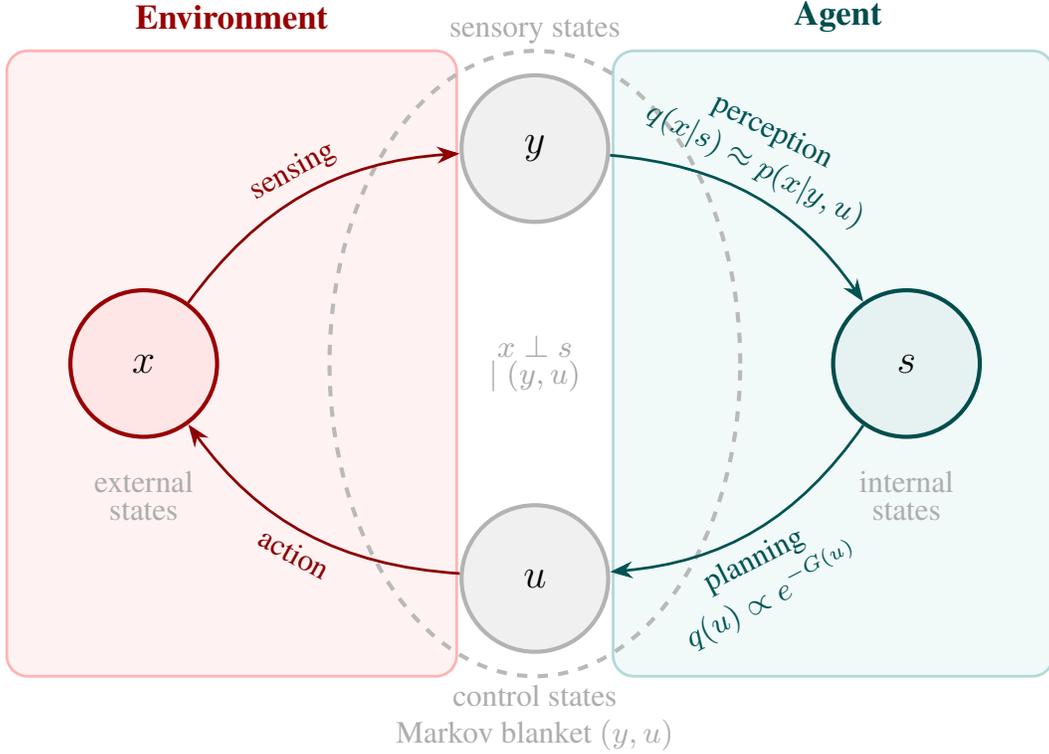
\begin{figure}[t]
\centering
\scalebox{1.3}{%
\begin{tikzpicture}[>=Stealth]

\fill[red!5,  rounded corners=6pt] (-5.4,-3.2) rectangle (-0.8, 3.2);
\draw[red!30, rounded corners=6pt, thick] (-5.4,-3.2) rectangle (-0.8, 3.2);
\node[font=\normalsize\bfseries, red!60!black] at (-3.1, 3.55) {Environment};

\fill[teal!5,  rounded corners=6pt] (0.8,-3.2) rectangle (5.3, 3.2);
\draw[teal!30, rounded corners=6pt, thick] (0.8,-3.2) rectangle (5.3, 3.2);
\node[font=\normalsize\bfseries, teal!60!black] at (3.1, 3.55) {Agent};

\draw[dashed, gray!55, line width=1.2pt] (0,0) ellipse (2.1cm and 3.2cm);
\node[font=\small, gray!65] at (0,-3.8) {Markov blanket $(y,u)$};

\node[circle, draw=red!60!black,  fill=red!10,  very thick,
      minimum size=1.5cm, font=\large]  (x) at (-4.0, 0) {$x$};
\node[font=\small, gray!65, align=center] at (-4.0,-1.35) {external\\[-2pt]states};

\node[circle, draw=teal!60!black, fill=teal!10, very thick,
      minimum size=1.5cm, font=\large]  (s) at (3.8, 0)  {$s$};
\node[font=\small, gray!65, align=center] at (3.8,-1.35) {internal\\[-2pt]states};

\node[circle, draw=gray!60, fill=gray!12, very thick,
      minimum size=1.5cm, font=\large]  (y) at (0,  2.2) {$y$};
\node[font=\small, gray!65] at (0, 3.4) {sensory states};

\node[circle, draw=gray!60, fill=gray!12, very thick,
      minimum size=1.5cm, font=\large]  (u) at (0, -2.2) {$u$};
\node[font=\small, gray!65] at (0,-3.4) {control states};

\node[font=\small, gray!70, align=center] at (0, 0)
    {$x \perp s$\\[-2pt]$\mid (y,u)$};

\draw[->, thick, red!55!black]
    (x) to[bend left=25]
    node[above, font=\small, sloped, midway] {sensing} (y);

\draw[->, thick, teal!65!black]
    (y) to[bend left=25]
    node[above, font=\small, sloped, align=center, pos=0.45]
        {perception\\[1pt]$q(x|s)\approx p(x|y,u)$} (s);

\draw[->, thick, teal!65!black]
    (s) to[bend left=25]
    node[below, font=\small, sloped, align=center, pos=0.55]
        {planning\\[1pt]$q(u)\propto e^{-G(u)}$} (u);

\draw[->, thick, red!55!black]
    (u) to[bend left=25]
    node[below, font=\small, sloped, midway] {action} (x);

\end{tikzpicture}%
}
\caption{Agent--environment interaction under the Free Energy Principle. States are partitioned into external $x$ (environment), sensory $y$, control $u$, and internal $s$ (agent). Sensory and control states jointly form the \emph{Markov blanket}, satisfying $x \perp s \mid (y,u)$. The perception--action loop runs clockwise: external states generate sensory inputs; the agent infers a posterior $q(x|s) \approx p(x|y,u)$ over external states (perception); internal states plan by forming a posterior $q(u) \propto e^{-G(u)}$ over control states, where $G(u)$ is the Expected Free Energy \eqref{eq:EFE-policy-posterior}; and the sampled action modifies the environment.}
\label{fig:agent-environment}
\end{figure}

Fig.~\ref{fig:agent-environment} illustrates this state partition and the resulting perception--action loop. This circular coupling between perception and action distinguishes physical AI agents from ``offline'' AI systems that operate without embodied interaction with their environment.

In this section, we briefly summarize the Free Energy Principle (FEP). Core references on its derivation include \citep{fristonfree2019,fristonfree2022}. The presentation in Section~\ref{sec:FEP} follows a physics-based route, starting from non-equilibrium steady-state dynamics. From an engineering perspective, the FEP (and its associated process, active inference) can be understood as a full commitment to VFE minimization as the unifying computational principle for physical AI agents interacting with their environment. Readers who prefer to take the FEP as given may skip directly to Section~\ref{sec:state-inference}, where the engineering development resumes.

\subsection{The Free Energy Principle: Dynamics of Systems that Maintain Their Identity}\label{sec:FEP}

The starting point of the FEP is the observation that many natural systems consume energy to self-organize and thereby maintain their structural and functional integrity over time \citep{fristonlife2013}. Living organisms are typical examples of such self-organizing systems. The FEP formalizes this intuition through the following assumptions \citep{fristonlife2013,fristonparcels2020,fristonfree2022,fristonpath2023}:

\begin{itemize}
    \item[\textbf{F1}] \label{fep:1} \textit{Langevin dynamics.}
    The combined states of a system and its environment evolve according to a stochastic
    differential equation, known as a Langevin process,
    \begin{equation}\label{eq:langevin}
        \dot{z} = f(z) + \omega \,, \quad \omega \sim \mathcal{N}(0,2\Gamma)\,,
    \end{equation}
    where $z$ denotes the collection of all system states,
    $\dot{z} \triangleq \mathrm{d}z/\mathrm{d}t$ is the time derivative,
    $f(z)$ is a \emph{deterministic flow} field, and $\omega$ is a Gaussian noise process.

    \item[\textbf{F2}] \label{fep:2} \textit{Non-equilibrium steady state.}
    The system admits a Non-Equilibrium Steady-State (NESS) density
    \begin{equation}\label{eq:steady-state}
        p(z)\,.
    \end{equation}
    Although the system is not at thermodynamic equilibrium, its probability density remains
    stationary over time. The system thereby occupies a restricted region of state space over
    extended periods~\citep{fristonlife2013} and maintains its recognizable structure.

    \item[\textbf{F3}] \label{fep:3} \textit{Markov blanket.}
    The system admits a partition of states $z = (y,x,u,s)$, where $y$ are sensory states,
    $x$ are external (environmental) states, $u$ are control (active) states, and $s$ are
    internal (agent) states. The sensory and control states jointly form a Markov blanket
    that statistically separates internal and external states~\citep{fristonlife2013,fristonparcels2020}, such that
    \begin{equation}
        x \perp s \,|\, (y,u).
    \end{equation}
    This identifies one part of the system as an \emph{agent} within the larger dynamical
    system and ensures that internal states are not directly coupled to external states.
\end{itemize}

The derivation of the core FEP result proceeds in three steps (see \citep[Part~One, Sections~1--2]{fristonfree2019} for the full treatment).

First, assumptions F1 and F2 together imply that the deterministic flow $f(z)$ admits a Helmholtz decomposition $f(z) = (\Gamma + Q)\nabla \log p(z)$ \citep[Eq.~(1.7)]{fristonfree2019}, where $\Gamma$ (defined in \eqref{eq:langevin}) is a symmetric (dissipative) diffusion matrix and $Q$ is an antisymmetric (solenoidal) matrix. The dissipative component $\Gamma \nabla \log p(z)$ drives the system toward regions of high steady-state probability, while the solenoidal component $Q \nabla \log p(z)$ circulates probability without changing the density.

Second, the Markov blanket condition (F3) ensures that the autonomous states $(s,u)$ are not directly coupled to external states $x$, so the autonomous components of the flow depend on $x$ only through the blanket states $(y,u)$ \citep[Eq.~(1.20), Marginal Flow Lemma]{fristonfree2019}.

Third, restricting the Helmholtz decomposition to the autonomous directions and marginalizing over $x$ yields a flow whose dissipative part is proportional to $\nabla_{(s,u)} \log p(y,u,s)$, giving \citep[Eq.~(8.3)]{fristonfree2019}, \citep[Eq.~(23)]{fristonfree2022}
\begin{subequations}\label{eq:auto-states-dynamics}
    \begin{align}
    \dot{s} &\propto \nabla_s \log p(y,u,s) \label{eq:internal-states-dynamics} \\
    \dot{u} &\propto \nabla_u \log p(y,u,s) \label{eq:control-states-dynamics}
\end{align}
\end{subequations}

If we define $\alpha = (s,u)$ as the autonomous states, then \eqref{eq:auto-states-dynamics} can also be written as
\begin{equation}
    \dot{\alpha} \propto  -\nabla_\alpha \big( \underbrace{-\log p(\alpha,y)}_{\text{surprisal}} \big)  \, ,
\end{equation}
which shows an AIF process can be interpreted as a dynamical system whose autonomous states evolve so as to minimize surprisal. Whenever the system drifts toward regions of lower steady-state probability, the dynamics \eqref{eq:auto-states-dynamics} tends to push it back toward more probable regions, thereby maintaining the agent's recognizable form. 

The dynamics in \eqref{eq:auto-states-dynamics} constitute the core mathematical result of the FEP under assumptions F1--F3. Under an additional variational interpretation, these dynamics may be read as an \emph{Active Inference} (AIF) process, in which internal states encode beliefs about external states and control states act so that sensations match predictions. We begin by showing how \eqref{eq:internal-states-dynamics} corresponds to VFE minimization over the internal states $s$, then extend this perspective to the control process.

\subsection{Internal State Estimation as VFE Minimization}\label{sec:state-inference}

Because internal states $s$ are coupled to but statistically insulated from external states $x$ by the Markov blanket (assumption F3), and because they follow a gradient flow on the steady-state density, their dynamics can be \emph{interpreted} as encoding the parameters of a conditional density over external states. In particular, we can define a map
\[
s \;\mapsto\; q(x|s),
\]
where $q(x|s)$ is a probability density over external states $x$, parameterized by the internal states $s$. This interpretive step is not forced by the dynamics themselves, which are fully specified by \eqref{eq:auto-states-dynamics}, but it is what enables the variational inference perspective that the remainder of this paper develops. The motivation is that internal states are continuously shaped by sensory input that itself depends on external states, so they necessarily accumulate statistical information about the environment. The map $s \mapsto q(x|s)$ makes this relationship explicit by interpreting the internal states as sufficient statistics of a belief distribution over external states. For instance, a Gaussian parameterization
\begin{equation}\label{eq:gaussian-q}
    q(x|s) = \mathcal{N}\!\left(x \,|\, \mu(s),\, \Sigma(s)\right)
\end{equation}
reads the internal states as encoding a mean $\mu(s)$ (best estimate of the world) and a covariance $\Sigma(s)$ (uncertainty about that estimate). The map is not unique: different parameterizations correspond to different inference schemes, analogous to the choice of variational family $\mathcal{Q}$ in \eqref{eq:q-hat-argmin}. For synthetic agents, this choice is a design decision. For a given density $q(x|s)$, we can then define an associated VFE functional\footnote{Formally, $F[q]$ in \eqref{eq:fep-vfe-def} depends on $y$ and $u$, and could therefore be written as $F[q](y,u)$. However, in the context of state updating, $y$ and $u$ represent \emph{past} observations and actions and are therefore treated as fixed. Consequently, $F[q]$ reduces to a function of beliefs over internal states $s$ alone.}
\begin{subequations}\label{eq:fep-vfe-def}
\begin{align}
    F[q] &=
    \int q(x|s)
    \log \frac{q(x|s)}{p(x,y,u)}
    \mathrm{d}x    \label{eq:fep-vfe-def-a} \\
    &= \underbrace{\int q(x|s)
    \log \frac{q(x|s)}{p(x|y,u)} \mathrm{d}x}_{\text{KL divergence} \geq 0} - \log p(y,u) \label{eq:fep-vfe-def-b}
\end{align}    
\end{subequations}
that aligns with the interpretation that $q(x|s)$ should approximate the Bayesian posterior $p(x|y,u)$. The denominator $p(x, y, u)$ in \eqref{eq:fep-vfe-def-a} is the generative model over the latent variable being inferred ($x$) and the observed blanket states ($y, u$).\footnote{For clarity, we distinguish three probability densities in this section. 
The density $p(z)=p(y,x,u,s)$ in \eqref{eq:steady-state} denotes the nonequilibrium 
steady-state (NESS) density of the coupled agent–environment system. 
The density $p(y,u,s)$ in \eqref{eq:auto-states-dynamics} is the marginal of this NESS 
density over the agent’s states. Finally, $p(y,x,u)$ in \eqref{eq:fep-vfe-def} denotes the agent’s \emph{generative model} of environmental dynamics.}  In this way, the steady-state gradient flow \eqref{eq:auto-states-dynamics} can be equivalently expressed as VFE minimization when internal states parameterize $q(x|s)$ \citep{fristonfree2022,fristonpath2023}. This process corresponds to perception: estimating the external causes $x$ of sensory inputs $y$. 

\subsection{Planning and Control as VFE Minimization}\label{sec:control}

Since actions influence future observations, inference over control states $u$ must be based on a rollout of the generative model to the future. In the following, $y$, $x$, and $u$ refer to sequences of future states, e.g., $u = (u_{t+1},u_{t+2},\ldots,u_{t+T})$. 

In the FEP literature, a sequence of future control states $u$ is called a \emph{policy} and inferring beliefs $q(u)$ over policies is called \emph{planning} \citep[Chapter~6]{parractive2022}. An \emph{action} involves an act of sampling the next control state from the marginal $q(u_{t+1})$ and applying the selected action to the environment. The particular sampling process, e.g., selecting the mode or drawing a random sample from $q(u_{t+1})$ depends on the specification of the actuator.   

The goal of this section is to show that planning and control in an AIF agent can be interpreted as variational inference over beliefs about policies $q(u)$ under a suitable generative model.

We start by considering a \emph{predictive} model rollout $p'(y,x,u)$ for environmental dynamics. Assume that the agent holds some  additional prior beliefs about the future. Firstly, we assume \emph{preference} priors $\hat{p}(x)$ that encode desired future states. Secondly, we assume \emph{epistemic} (``information-seeking'') priors beliefs 
\begin{subequations}\label{eq:epistemic-priors}
\begin{align}
    \tilde{p}(u) &= \exp\big(\mathbb{H}[q(x|u)]\big)\,, \label{eq:u-epistemic-prior}\\
    \tilde{p}(x) &= \exp\big(-\mathbb{H}[q(y|x)]\big)\,, \label{eq:x-epistemic-prior}
\end{align}\end{subequations}

where $\mathbb{H}[q] = \mathbb{E}_q[-\log q]$ is the Shannon entropy of $q$. The role of these epistemic priors will be revealed shortly.

The agent's beliefs about the future are then represented by the generative model \citep{vriesexpected2025,nuijtenmessage2026}

\begin{equation}\label{eq:generative-model-for-planning}
   f(y,x,u) \propto \underbrace{p'(y,x,u)}_{\text{predictive}}\underbrace{\hat{p}(x)}_{\text{preference}}\underbrace{\tilde{p}(x)\tilde{p}(u)}_{\text{epistemic}}\,.
\end{equation}

We then define and work out the corresponding VFE as (see Appendix~\ref{app:vfe-control-derivation} for the derivation)
\begin{subequations}\label{eq:VFE-for-control}
\begin{align}
    F[q] &= \mathbb{E}_{q(y,x,u)}\Big[ \log \frac{q(y,x,u)}{f(y,x,u)}\Big]  \\
    &=\mathbb{E}_{q(u)}\Bigg[ \log \frac{q(u)}{\exp\big(-G(u) -C(u) - P(u)\big)}\Bigg] \qquad \text{if }\eqref{eq:epistemic-priors} \text{ holds}\,,\label{eq:Fqu-as-KLD}
\end{align}
\end{subequations}
where 
\begin{subequations}\label{eq:GuCuPu-def}
\begin{align}
    G(u) &=
    \underbrace{\mathbb{E}_{q(y,x|u)}\left[ \log \frac{q(x|u)}{\hat{p}(x)} \right]}_{\text{risk}}
    +
    \underbrace{\mathbb{E}_{q(y,x|u)}\left[ \log \frac{1}{q(y|x)} \right]}_{\text{ambiguity}}  &&\text{(expected free energy)}\label{eq:Gu-def} \\
    C(u) &= \mathbb{E}_{q(y,x|u)}\Big[ \log \frac{q(y,x|u)}{p'(y,x|u)} \Big]  &&\text{(complexity)}\label{eq:Cu-def} \\
    P(u) &= -\log p'(u)\,. &&\text{(empirical policy prior)}\label{eq:Pu-def}
\end{align}
\end{subequations}

Since \eqref{eq:Fqu-as-KLD} is a KL divergence, minimization of $F[q]$ leads to 
\begin{align}\label{eq:EFE-policy-posterior}
  q^*(u) = \arg\min_q F[q] = \sigma\big(-G(u) -C(u) - P(u)\big)
\end{align}
where
\begin{equation}
    \sigma(a)_k = \frac{\exp(a_k)}{\sum_{k'} \exp(a_{k'})}
\end{equation}
is a normalized exponential (``softmax'') function. Next, we discuss these results in more detail. 

The first observation is that minimizing $F[q]$ in \eqref{eq:VFE-for-control} yields a posterior belief $q^*(u)$ over policies \eqref{eq:EFE-policy-posterior} that is influenced by three forces: $G(u)$, $C(u)$, and $P(u)$, as defined in \eqref{eq:GuCuPu-def}. The most important conclusion is that, in an AIF agent, updating beliefs about policies can be realized by standard VFE minimization in an appropriate generative model. In other words, the idea of planning-as-inference is fully realized in AIF agents \citep{attiasplanning2003,botvinickplanning2012a}. 

\emph{Expected Free Energy} $G(u)$: The term $G(u)$ in \eqref{eq:Gu-def} is known as the \emph{Expected Free Energy} (EFE) cost function for a given policy $u$ \citep[Appendix D]{dacostaactive2024}. The EFE evaluates the predicted consequences of a policy by trading off \emph{risk} and \emph{ambiguity}. The risk term penalizes deviations of predicted future states $q(x | u)$ from preferred states $\hat{p}(x)$, thereby promoting goal-directed behavior. The ambiguity term, which evaluates to an expected conditional entropy $\mathbb{E}_{q(x|u)}\bigl[ \mathbb{H}\left[q(y | x)\right]\bigr]$ of observations $y$ given a policy $u$, penalizes policies that are expected to generate uninformative observations, thus giving rise to information-seeking behavior. Both exploitation and exploration therefore emerge from minimizing a single objective. Models that accurately predict future outcomes tend to favor goal-directed policies, whereas models with poor predictive accuracy naturally favor information-seeking policies. The emergence of a balanced exploration–exploitation trade-off is a key advantage over standard reinforcement learning and decision-theoretic approaches, where the objective is typically formulated as a value (or reward) function of states rather than as a function of beliefs about states. As will be discussed in Section~\ref{sec:exploratory-behavior}, the EFE follows naturally from first principles in nested AIF agents. 

The specific form of the epistemic priors in \eqref{eq:epistemic-priors} is what makes the clean decomposition \eqref{eq:Fqu-as-KLD} possible (see Appendix~\ref{app:vfe-control-derivation}). Their role can be understood as follows. The prior $\tilde{p}(u) = \exp\big(\mathbb{H}[q(x|u)]\big)$ in \eqref{eq:u-epistemic-prior} assigns high probability to policies $u$ under which the predicted future states $x$ have high entropy, thereby biasing the agent toward maintaining flexibility and postponing premature commitment. The prior $\tilde{p}(x) = \exp\big(-\mathbb{H}[q(y|x)]\big)$ in \eqref{eq:x-epistemic-prior} assigns high probability to future states $x$ from which observations $y$ are highly informative, i.e., states where the conditional entropy $\mathbb{H}[q(y|x)]$ is low. Together, these two priors induce the ambiguity-minimizing component of the EFE: the agent preferentially selects policies that lead to informative observations about future states.

\emph{Complexity} $C(u)$: The term $C(u)$ in \eqref{eq:Cu-def} is a \emph{complexity} term that penalizes policies based on a predicted future that deviates from predictions by $p'(y,x,u)$. This term plays an analogous role as the complexity term in the complexity-minus-accuracy decomposition of the VFE that we derived in \eqref{eq:CA-decomp}. Since $p'(y,x,u)$ reflects a learned model from past observations, deviating from its predictions is potentially dangerous. The complexity cost thus acts as a regularizer in \eqref{eq:VFE-for-control}, anchoring the agent's policy selection to its prior experience and preventing overly speculative action plans. 

\emph{Empirical Prior} $P(u)$: $P(u)$ in \eqref{eq:Pu-def} denotes the prior belief over policies and captures structural or habitual biases that favor certain policies independent of their predicted outcomes.

In summary, through continual minimization of \eqref{eq:VFE-for-control}, the updating of beliefs about policies is guided by balancing three objectives: the expected free energy $G(u)$, which trades off goal-directed and information-seeking behavior; the complexity cost $C(u)$, which anchors policy selection to prior experience; and the policy prior $P(u)$, which encodes any additional preferences or constraints over admissible policies. 

The crucial conclusion of this section is that policy selection by EFE minimization does not introduce a new principle; it arises from standard variational inference under the modeling commitments described above.

\subsection{Nested Active Inference Agents}\label{sec:nested-aif}

Section~\ref{sec:FEP} established that systems satisfying assumptions F1--F3 admit an active-inference interpretation under the variational reading introduced above. A natural follow-up question is whether the internal states $s$ of an AIF agent can themselves be composed of lower-level AIF agents, and if so, under what conditions. This question matters practically: if the answer is yes, then one can build large-scale intelligent systems by composing simpler AIF agents, without ever leaving the VFE-minimization framework. This would offer significant engineering benefits, as the engineering effort can then focus solely on efficient VFE minimization, which can be realized in a way that is very suitable for dealing with typical physical AI constraints (to be discussed in Section~\ref{sec:factor-graphs}). The answer, developed in~\citet{fristonfree2019,fristonparcels2020,hipólitomarkov2021,fagerholmneural2021}, is that, under suitable structural and coarse-graining conditions, a collection of coupled AIF agents can itself be described as a higher-level AIF agent.

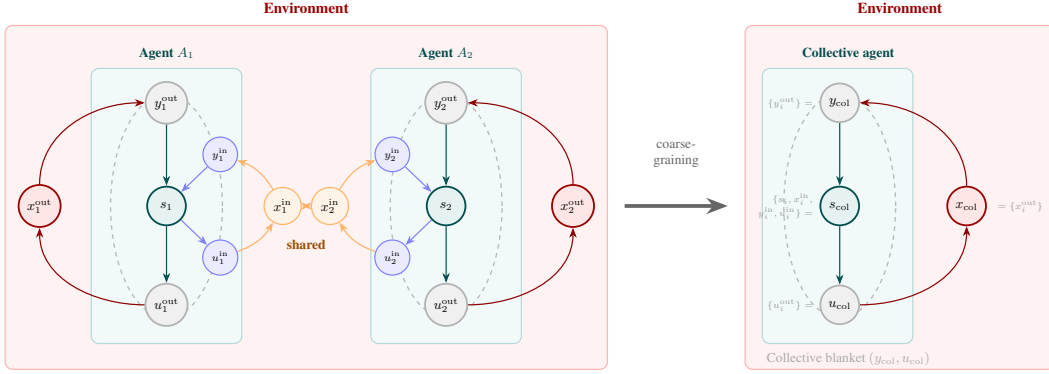
\begin{figure}[t]
\centering
\resizebox{\columnwidth}{!}{%
\begin{tikzpicture}[>=Stealth, every node/.style={font=\small}]


\fill[red!5, rounded corners=5pt] (-7.0, -5.4) rectangle (7.0, 2.6);
\draw[red!30, rounded corners=5pt, thick] (-7.0, -5.4) rectangle (7.0, 2.6);
\node[font=\normalsize\bfseries, red!60!black] at (0, 3.0)
    {Environment};

\node[circle, draw=red!60!black, fill=red!10, very thick,
      minimum size=0.9cm, font=\footnotesize] (x1out) at (-6.2, -1.6) {$x_1^{\mathrm{out}}$};
\node[circle, draw=red!60!black, fill=red!10, very thick,
      minimum size=0.9cm, font=\footnotesize] (x2out) at (6.2, -1.6) {$x_2^{\mathrm{out}}$};


\node[circle, draw=orange!60, fill=orange!10, thick,
      minimum size=0.8cm, font=\footnotesize] (x1in) at (-0.55, -1.6) {$x_1^{\mathrm{in}}$};
\node[circle, draw=orange!60, fill=orange!10, thick,
      minimum size=0.8cm, font=\footnotesize] (x2in) at (0.55, -1.6) {$x_2^{\mathrm{in}}$};

\node[font=\footnotesize\bfseries, orange!60!black] at (0, -2.5) {shared};

\draw[<->, thick, orange!60] (x1in) -- (x2in);

\fill[teal!5, rounded corners=5pt] (-5.0, -4.8) rectangle (-1.5, 1.6);
\draw[teal!30, rounded corners=5pt, thick] (-5.0, -4.8) rectangle (-1.5, 1.6);
\node[font=\small\bfseries, teal!60!black] at (-3.25, 1.95) {Agent $A_1$};

\draw[dashed, gray!55, line width=0.9pt] (-3.25, -1.6) ellipse (1.3cm and 2.6cm);

\node[circle, draw=gray!60, fill=gray!12, very thick,
      minimum size=0.9cm, font=\footnotesize] (y1out) at (-3.25, 0.8) {$y_1^{\mathrm{out}}$};

\node[circle, draw=teal!60!black, fill=teal!10, very thick,
      minimum size=0.9cm] (s1) at (-3.25, -1.6) {$s_1$};

\node[circle, draw=gray!60, fill=gray!12, very thick,
      minimum size=0.9cm, font=\footnotesize] (u1out) at (-3.25, -3.9) {$u_1^{\mathrm{out}}$};

\node[circle, draw=blue!50, fill=blue!8, thick,
      minimum size=0.7cm, font=\scriptsize] (y1in) at (-2.0, -0.4) {$y_1^{\mathrm{in}}$};
\node[circle, draw=blue!50, fill=blue!8, thick,
      minimum size=0.7cm, font=\scriptsize] (u1in) at (-2.0, -2.8) {$u_1^{\mathrm{in}}$};

\draw[->, thick, teal!65!black] (y1out) -- (s1);
\draw[->, thick, teal!65!black] (s1) -- (u1out);
\draw[->, thick, blue!50] (y1in) -- (s1);
\draw[->, thick, blue!50] (s1) -- (u1in);

\fill[teal!5, rounded corners=5pt] (1.5, -4.8) rectangle (5.0, 1.6);
\draw[teal!30, rounded corners=5pt, thick] (1.5, -4.8) rectangle (5.0, 1.6);
\node[font=\small\bfseries, teal!60!black] at (3.25, 1.95) {Agent $A_2$};

\draw[dashed, gray!55, line width=0.9pt] (3.25, -1.6) ellipse (1.3cm and 2.6cm);

\node[circle, draw=gray!60, fill=gray!12, very thick,
      minimum size=0.9cm, font=\footnotesize] (y2out) at (3.25, 0.8) {$y_2^{\mathrm{out}}$};

\node[circle, draw=teal!60!black, fill=teal!10, very thick,
      minimum size=0.9cm] (s2) at (3.25, -1.6) {$s_2$};

\node[circle, draw=gray!60, fill=gray!12, very thick,
      minimum size=0.9cm, font=\footnotesize] (u2out) at (3.25, -3.9) {$u_2^{\mathrm{out}}$};

\node[circle, draw=blue!50, fill=blue!8, thick,
      minimum size=0.7cm, font=\scriptsize] (y2in) at (2.0, -0.4) {$y_2^{\mathrm{in}}$};
\node[circle, draw=blue!50, fill=blue!8, thick,
      minimum size=0.7cm, font=\scriptsize] (u2in) at (2.0, -2.8) {$u_2^{\mathrm{in}}$};

\draw[->, thick, teal!65!black] (y2out) -- (s2);
\draw[->, thick, teal!65!black] (s2) -- (u2out);
\draw[->, thick, blue!50] (y2in) -- (s2);
\draw[->, thick, blue!50] (s2) -- (u2in);

\draw[->, thick, orange!60] (u1in) to[bend right=20] (x1in);
\draw[->, thick, orange!60] (x2in) to[bend left=20] (y2in);
\draw[->, thick, orange!60] (u2in) to[bend left=20] (x2in);
\draw[->, thick, orange!60] (x1in) to[bend right=20] (y1in);

\draw[->, thick, red!55!black] (x1out) to[out=90, in=180] (y1out);
\draw[->, thick, red!55!black] (u1out) to[out=180, in=270] (x1out);
\draw[->, thick, red!55!black] (x2out) to[out=90, in=0] (y2out);
\draw[->, thick, red!55!black] (u2out) to[out=0, in=270] (x2out);

\node[font=\small, align=center, black!60] at (8.6, -0.4) {coarse-\\graining};
\draw[->, line width=2.5pt, black!60] (7.4, -1.6) -- (9.8, -1.6);


\fill[red!5, rounded corners=5pt] (10.2, -5.4) rectangle (17.4, 2.6);
\draw[red!30, rounded corners=5pt, thick] (10.2, -5.4) rectangle (17.4, 2.6);
\node[font=\normalsize\bfseries, red!60!black] at (13.8, 3.0)
    {Environment};

\node[circle, draw=red!60!black, fill=red!10, very thick,
      minimum size=0.9cm, font=\footnotesize] (xcol) at (15.35, -1.6) {$x_{\mathrm{col}}$};
\node[font=\tiny, gray!80, anchor=west] at (15.95, -1.6)
    {$= \{x_i^{\mathrm{out}}\}$};

\fill[teal!5, rounded corners=5pt] (10.6, -4.8) rectangle (14.1, 1.6);
\draw[teal!30, rounded corners=5pt, thick] (10.6, -4.8) rectangle (14.1, 1.6);
\node[font=\small\bfseries, teal!60!black] at (12.6, 1.95) {Collective agent};

\draw[dashed, gray!55, line width=0.9pt] (12.4, -1.6) ellipse (1.3cm and 2.6cm);
\node[font=\footnotesize, gray!65] at (12.6, -5.15)
    {Collective blanket $(y_{\mathrm{col}}, u_{\mathrm{col}})$};

\node[circle, draw=gray!60, fill=gray!12, very thick,
      minimum size=0.9cm, font=\footnotesize] (ycol) at (12.4, 0.8)
      {$y_{\mathrm{col}}$};
\node[font=\tiny, gray!80, anchor=east] at (11.9, 0.8)
    {$\{y_i^{\mathrm{out}}\} =$};

\node[circle, draw=teal!60!black, fill=teal!10, very thick,
      minimum size=0.9cm, font=\footnotesize] (scol) at (12.4, -1.6) {$s_{\mathrm{col}}$};
\node[font=\tiny, gray!80, anchor=east, align=right] at (11.9, -1.6)
    {$\{s_i, x_i^{\mathrm{in}},$\\$y_i^{\mathrm{in}}, u_i^{\mathrm{in}}\} =$};

\node[circle, draw=gray!60, fill=gray!12, very thick,
      minimum size=0.9cm, font=\footnotesize] (ucol) at (12.4, -3.9)
      {$u_{\mathrm{col}}$};
\node[font=\tiny, gray!80, anchor=east] at (11.9, -3.9)
    {$\{u_i^{\mathrm{out}}\} =$};

\draw[->, thick, teal!65!black] (ycol) -- (scol);
\draw[->, thick, teal!65!black] (scol) -- (ucol);

\draw[->, thick, red!55!black] (xcol) to[out=90, in=0] (ycol);
\draw[->, thick, red!55!black] (ucol) to[out=0, in=270] (xcol);

\end{tikzpicture}%
}
\caption{Nested active inference: coarse-graining two coupled AIF agents into a collective AIF agent. \emph{Left:} Two agents $A_1$ and $A_2$, each with partition $(y_i, x_i, u_i, s_i)$. External states are split into outward-facing $x_i^{\mathrm{out}}$ (red), which interface with the environment beyond the ensemble, and inward-facing $x_i^{\mathrm{in}}$ (orange), which mediate inter-agent coupling. Similarly, blanket states are split into outward-facing $(y_i^{\mathrm{out}}, u_i^{\mathrm{out}})$ (gray) and inward-facing $(y_i^{\mathrm{in}}, u_i^{\mathrm{in}})$ (blue). Inter-agent influence follows the chain $s_i \to u_i^{\mathrm{in}} \to x_i^{\mathrm{in}} \leftrightarrow x_j^{\mathrm{in}} \to y_j^{\mathrm{in}} \to s_j$. \emph{Right:} Under conditions E1--E3, the ensemble admits a collective Markov-blanketed partition \eqref{eq:collective-partition}. Outward-facing blanket states form the collective blanket $(y_{\mathrm{col}}, u_{\mathrm{col}})$, outward-facing external states form $x_{\mathrm{col}}$, and all inward-facing states together with the individual internal states are absorbed into $s_{\mathrm{col}}$.}
\label{fig:nested-aif}
\end{figure}

Consider $N$ interacting AIF agents $\{A_i\}_{i=1}^{N}$, each admitting a 
Markov–blanketed partition $(y_i,x_i,u_i,s_i)$, where $y_i$ are sensory
states, $x_i$ external states, $u_i$ active states, and $s_i$ internal states. 
For each agent the pair $(y_i,u_i)$ forms its Markov blanket, and the 
autonomous states are $\alpha_i=(s_i,u_i)$.

When multiple agents interact, not all blanket states play the same role at 
the collective scale. We therefore distinguish between 
\emph{outward-facing} blanket states, which interface with the environment 
external to the ensemble, and \emph{inward-facing} blanket states, which 
mediate coupling between agents within the ensemble. Specifically, we write
\[
x_i = (x_i^{\mathrm{out}},x_i^{\mathrm{in}}),
\qquad
u_i = (u_i^{\mathrm{out}},u_i^{\mathrm{in}}), 
\qquad
y_i = (y_i^{\mathrm{out}},y_i^{\mathrm{in}}),
\]
where the superscripts denote outward- and inward-facing components. Here,
$x_i^{\mathrm{in}}$ denotes those components of agent $i$'s external states
that mediate coupling to other agents in the ensemble, while
$x_i^{\mathrm{out}}$ denotes those components that remain external to the
ensemble as a whole. Similarly, $y_i^{\mathrm{in}}$ and $u_i^{\mathrm{in}}$
are the sensory and active states through which agent $i$ observes and
influences other agents, while $y_i^{\mathrm{out}}$ and $u_i^{\mathrm{out}}$
face the external environment.

Under conditions E1--E3 below, the ensemble admits a coarse-grained
collective Markov-blanketed partition
\begin{equation}\label{eq:collective-partition}
\begin{aligned}
u_{\mathrm{col}} &= \{u_i^{\mathrm{out}}\}_{i=1}^{N}, &
y_{\mathrm{col}} &= \{y_i^{\mathrm{out}}\}_{i=1}^{N}, \\
s_{\mathrm{col}} &= \{s_i,\,x_i^{\mathrm{in}},\,u_i^{\mathrm{in}},\,y_i^{\mathrm{in}}\}_{i=1}^{N}, &
x_{\mathrm{col}} &= \{x_i^{\mathrm{out}}\}_{i=1}^{N}.
\end{aligned}
\end{equation}
Thus, outward-facing blanket states form the \emph{collective Markov blanket} 
$(y_{\mathrm{col}},u_{\mathrm{col}})$ that mediates interaction with the 
external environment $x_{\mathrm{col}}$. In contrast, inward-facing blanket 
states and the corresponding inter-agent external states
$x_i^{\mathrm{in}}$ are absorbed into the collective internal states, since
at the ensemble scale they mediate communication among agents rather than
coupling to the external world.

Under the following conditions, the ensemble $\{A_i\}_{i=1}^{N}$ can be 
described as an AIF agent at the collective scale.

\begin{itemize}

\item[\textbf{E1}] \emph{Blanket-mediated coupling.}  
Interactions between agents occur through their blanket states. 
In particular, cross-agent influence is mediated through shared
environmental states, following the chain
\[
s_i \rightarrow u_i^{\mathrm{in}} \rightarrow x_i^{\mathrm{in}} \leftrightarrow x_j^{\mathrm{in}} \rightarrow y_j^{\mathrm{in}} \rightarrow s_j,
\qquad i\neq j,
\]
rather than through a direct coupling between internal states.
This sparse coupling 
structure ensures that the outward blanket states jointly screen the 
collective internal states $s_{\mathrm{col}}$ from the collective external 
states $x_{\mathrm{col}}$, yielding a ``blanket of blankets''
\citep{kirchhoffmarkov2018,fristonparcels2020}.

\item[\textbf{E2}] \emph{Timescale separation.}  
Local inference dynamics within each agent operate on a faster timescale 
than the inter-agent coupling. Consequently, fast intra-agent fluctuations 
can be approximated by quasi-stationary averages, yielding a reduced 
description in terms of slow collective modes. This coarse-graining allows 
the ensemble dynamics to be expressed in terms of collective variables, to 
which the variational free-energy interpretation can subsequently be applied (if conditions 
E1 and E3 also hold)~\citep{fristonparcels2020,fristonfree2019}.

\item[\textbf{E3}] \emph{Collective steady state.}  
The coupled ensemble admits a non-equilibrium steady-state density 
$p(z_{\mathrm{col}})$ over the collective state 
$z_{\mathrm{col}}=(y_{\mathrm{col}},x_{\mathrm{col}},
u_{\mathrm{col}},s_{\mathrm{col}})$. This ensures that the Helmholtz 
decomposition applies at the collective scale and that the ensemble behaves 
as a coherent self-organizing system \citep{kirchhoffmarkov2018,
fristonlife2013}.

\end{itemize}

In summary, E1 ensures the existence of a collective Markov blanket, E2 enables coarse-graining that produces collective variables evolving on a slower timescale than the constituent agents, and E3 guarantees the existence of a steady-state density, allowing the Helmholtz decomposition to be applied at the collective scale. When E1--E3 hold, analogous to 
\eqref{eq:auto-states-dynamics}, the collective autonomous states
$\alpha_{\mathrm{col}}=(s_{\mathrm{col}},u_{\mathrm{col}})$ obey a gradient 
flow on the collective steady-state density,
\begin{equation}
\dot{\alpha}_{\mathrm{col}}
\;\propto\;
\nabla_{\alpha_{\mathrm{col}}}
\log p(y_{\mathrm{col}},u_{\mathrm{col}},s_{\mathrm{col}})\,,
\end{equation}
which admits the usual variational free energy
interpretation. The ensemble 
therefore behaves as an AIF agent at the collective scale without introducing any new principle beyond 
F1--F3, illustrating the 
nested or multi-scale nature of active inference systems~\citep{fristonparcels2020}.

\subsection{Exploratory Behavior Emerges from Nested AIF Agents}\label{sec:exploratory-behavior}

One important consequence of the coarse-graining operations (E2) over both 
time and space that accompany the nesting of active inference agents is 
that fast fluctuations tend to average out. As a result, larger AIF agents 
composed of multiple levels of nested sub-agents exhibit increasingly 
deterministic responses to environmental observations. This motivates the 
concept of a \emph{precise} agent: a macroscopic AIF agent whose collective 
autonomous states $\alpha_{\mathrm{col}}$ respond nearly deterministically 
to sensory inputs $y_{\mathrm{col}}$. In~\citet{dacostaactive2024,barpgeometric2022}, this regime is modeled by 
assuming that $p(\alpha | y,u)$ is effectively deterministic (i.e., a delta 
distribution). In~\citet{vriesexpected2025,nuijtenmessage2026}, also followed in Section~\ref{sec:control}, a similar 
effect is achieved by augmenting the generative model with specific epistemic prior 
beliefs~\eqref{eq:epistemic-priors}.

A deterministic response does not imply that the posterior beliefs 
$q(\alpha_{\mathrm{col}})$ are free of uncertainty. Rather, it means that 
any remaining uncertainty reflects incomplete knowledge of environmental 
dynamics, rather than stochasticity in the agent's own internal processing. Since the agent ``knows'' that its uncertainty pertains only to the environment, it can select actions that effectively query the environment to reduce this uncertainty. Formally, the variational free energy objective admits a decomposition in which policies are evaluated in terms of expected free energy (EFE) (Section~\ref{sec:control}). The resulting EFE objective 
naturally contains both goal-directed (risk) and information-seeking 
(ambiguity) terms, giving rise to exploratory behavior. In short, the 
characteristic balance between goal-directed and information-seeking 
behavior in macroscopic AIF agents is not an ad-hoc mechanism but an 
emergent property of nesting AIF agents: coarse-graining over fast internal 
fluctuations yields precise agents whose remaining uncertainty is purely 
epistemic, and epistemic uncertainty is precisely what drives the 
information-seeking component of the EFE.

The transition from stochastic microscopic dynamics to near-deterministic macroscopic behavior through successive coarse-graining has a well-known analogue in theoretical physics, where averaging over microscopic quantum fluctuations yields macroscopic dynamics governed by a variational (least-action) principle~\citep{feynmanquantum1965}. Friston draws an explicit analogy between this quantum-to-classical transition and the coarse-graining of nested AIF agents~\citep{fristonpath2023,fristonfree2022}.

\section{Realization: Factor Graphs and Reactive Message Passing}\label{sec:factor-graphs}

\subsection{Forney-style Factor Graphs}\label{sec:forney-factor-graphs}

A \emph{Forney-style factor graph} (FFG) is a
graphical representation of a factored joint probability distribution \citep{kschischangfactor2001,loeligerintroduction2004,loeligerfactor2007}. In an FFG,
square nodes represent \emph{factors} (local functions), edges represent \emph{variables},
and the graph encodes the factorization of the joint distribution as the product of those factors.
Each factor node is connected only to the edges that correspond to its arguments,
making the conditional independence structure of the model explicit.

For our purposes, the importance of FFGs is that they expose inference as a collection of strictly local computations, making them a natural computational substrate for distributed VFE minimization in physical AIF agents.

Consider a joint distribution over variables $x_1,\ldots,x_7$ that factorizes as\footnote{Here, $x$ denotes a generic variable and should not be confused with the previously introduced environmental state variable $x$.}
\begin{equation}\label{eq:fg-factorization}
f(x_1,\ldots,x_7)
= f_a(x_1)\,f_b(x_2)\,f_c(x_1,x_2,x_3)\,f_d(x_4)\,f_e(x_3,x_4,x_5)\,f_f(x_5,x_6,x_7)\,f_g(x_7)\,.
\end{equation}

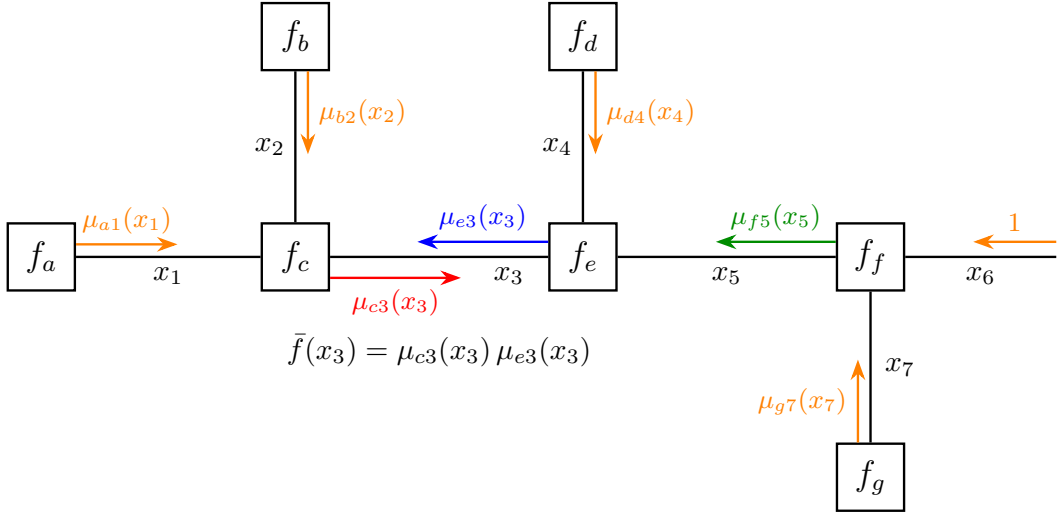
\begin{figure}[!t]
\centering
\resizebox{\columnwidth}{!}{%
\begin{tikzpicture}[
    factor/.style={draw, rectangle, thick, minimum width=0.8cm, minimum height=0.8cm, font=\large},
    edgeline/.style={thick},
    msgarrow/.style={-{Stealth[length=7pt,width=5pt]}, thick},
]

\node[factor] (fa) {$f_a$};
\node[factor, right=2.2cm of fa] (fc) {$f_c$};
\node[factor, above=1.8cm of fc] (fb) {$f_b$};
\node[factor, right=2.6cm of fc] (fe) {$f_e$};
\node[factor, above=1.8cm of fe] (fd) {$f_d$};
\node[factor, right=2.6cm of fe] (ff) {$f_f$};
\node[factor, below=1.8cm of ff] (fg) {$f_g$};

\draw[edgeline] (fa.east) -- (fc.west);
\node[below] at ($(fa.east)!0.5!(fc.west)$) {$x_1$};

\draw[edgeline] (fb.south) -- (fc.north);
\node[left] at ($(fb.south)!0.5!(fc.north)$) {$x_2$};

\draw[edgeline] (fc.east) -- (fe.west);
\node[below] at ($(fc.east)!0.82!(fe.west)$) {$x_3$};

\draw[edgeline] (fd.south) -- (fe.north);
\node[left] at ($(fd.south)!0.5!(fe.north)$) {$x_4$};

\draw[edgeline] (fe.east) -- (ff.west);
\node[below] at ($(fe.east)!0.5!(ff.west)$) {$x_5$};

\coordinate (x6end) at ($(ff.east)+(1.8,0)$);
\draw[edgeline] (ff.east) -- (x6end);
\node[below] at ($(ff.east)!0.5!(x6end)$) {$x_6$};

\draw[edgeline] (ff.south) -- (fg.north);
\node[right] at ($(ff.south)!0.5!(fg.north)+(0.05,0)$) {$x_7$};


\draw[msgarrow, orange]
    ($(fa.east)+(0,0.15)$) -- ($(fa.east)!0.55!(fc.west)+(0,0.15)$)
    node[above, midway, font=\small, orange] {$\mu_{a1}(x_1)$};

\draw[msgarrow, orange]
    ($(fb.south)+(0.15,0)$) -- ($(fb.south)!0.55!(fc.north)+(0.15,0)$)
    node[right, midway, font=\small, orange] {$\mu_{b2}(x_2)$};

\draw[msgarrow, blue]
    ($(fe.west)+(0,0.18)$) -- ($(fe.west)!0.60!(fc.east)+(0,0.18)$)
    node[above, midway, font=\small, blue] {$\mu_{e3}(x_3)$};

\draw[msgarrow, red]
    ($(fc.east)+(0,-0.25)$) -- ($(fc.east)!0.60!(fe.west)+(0,-0.25)$)
    node[below, midway, font=\small, red] {$\mu_{c3}(x_3)$};

\draw[msgarrow, orange]
    ($(fd.south)+(0.15,0)$) -- ($(fd.south)!0.55!(fe.north)+(0.15,0)$)
    node[right, midway, font=\small, orange] {$\mu_{d4}(x_4)$};

\draw[msgarrow, green!55!black]
    ($(ff.west)+(0,0.18)$) -- ($(ff.west)!0.55!(fe.east)+(0,0.18)$)
    node[above, midway, font=\small, green!55!black] {$\mu_{f5}(x_5)$};

\draw[msgarrow, orange]
    ($(x6end)+(0,0.18)$) -- ($(ff.east)!0.45!(x6end)+(0,0.18)$)
    node[above, midway, font=\small, orange] {$1$};

\draw[msgarrow, orange]
    ($(fg.north)+(-0.15,0)$) -- ($(fg.north)!0.55!(ff.south)+(-0.15,0)$)
    node[left, midway, font=\small, orange] {$\mu_{g7}(x_7)$};

\node[font=\normalsize, anchor=north] at ($(fc.east)!0.5!(fe.west)+(0,-0.8)$)
    {$\bar{f}(x_3) = \mu_{c3}(x_3)\,\mu_{e3}(x_3)$};

\end{tikzpicture}%
}
\caption{A Forney-style factor graph with messages $\mu_{ai}(x_i)$, denoting the message from node $f_a$ to edge $x_i$, and the resulting marginal $\bar{f}(x_3)$ as defined in \eqref{eq:fg-marginal-product}.}
\label{fig:forney_fg}
\end{figure}

Fig.~\ref{fig:forney_fg} shows the corresponding FFG. A central inference task is computing a marginal, e.g., the marginal of $x_3$:\footnote{We use discrete sums here for illustration; in the continuous case, sums are replaced by integrals.}
\begin{align}\label{eq:fg-marginal-naive}
\bar{f}(x_3) &= \sum_{x_1}\sum_{x_2}\sum_{x_4}\sum_{x_5}\sum_{x_6}\sum_{x_7
} f(x_1,\ldots,x_7) \,.
\end{align}
If each variable has $10$ possible values, then direct evaluation scales on the
order of $10^6$ arithmetic operations. However, substituting the factorization~\eqref{eq:fg-factorization} and applying the
\emph{distributive law} (moving integrals inward past factors that do not depend on
the integration variable) transforms \eqref{eq:fg-marginal-naive} into the following product-of-sums, 

\begin{align}\label{eq:fg-marginal-split}
\bar{f}(x_3) =
  &\underbrace{ \Bigg( \sum_{x_1,x_2} \underbrace{f_a(x_1)}_{\mu_{a1}(x_1)}\, \underbrace{f_b(x_2)}_{\mu_{b2}(x_2)}\,f_c(x_1,x_2,x_3)\Bigg) }_{\mu_{c3}(x_3)} \notag \\
  &\quad\underbrace{ \cdot\Bigg( \sum_{x_4,x_5} \underbrace{f_d(x_4)}_{\mu_{d4}(x_4)}\,f_e(x_3,x_4,x_5) \cdot \underbrace{ \big( \sum_{x_6,x_7} f_f(x_5,x_6,x_7)\,\underbrace{f_g(x_7)}_{\mu_{g7}(x_7)}\big) }_{\mu_{f5}(x_5)} \Bigg) }_{\mu_{e3}(x_3)}\,,
\end{align}
which executes with only a few hundred local operations. Clearly, the computational gains obtained by exploiting this distributive structure are substantial.

The intermediate results in \eqref{eq:fg-marginal-split} admit a graphical interpretation as messages passed along the factor graph. For example, the intermediate result
\begin{equation}\label{eq:mu-c3-sum-product}
    \mu_{c3}(x_3) = \sum_{x_1,x_2} \mu_{a1}(x_1) \mu_{b2}(x_2) f_c(x_1,x_2,x_3)
\end{equation}
can be interpreted as an \emph{outgoing} message from factor node $f_c$ to the variable edge $x_3$. The computation rule is: multiply incoming messages with the node function and integrate out the variables on the incoming edges. This is the \emph{sum-product rule}. The Bayesian inference procedure in \eqref{eq:fg-marginal-split}, in which posterior marginals are obtained by propagating such messages through the FFG, is called the \emph{sum-product algorithm}. 

Importantly, message \eqref{eq:mu-c3-sum-product} can be computed locally at node $f_c$ using only the \emph{incoming} messages $\mu_{a1}(x_1)=f_a(x_1)$ and $\mu_{b2}(x_2)=f_b(x_2)$. The marginal
\begin{equation}\label{eq:fg-marginal-product}
\bar{f}(x_3)=\mu_{c3}(x_3)\,\mu_{e3}(x_3)
\end{equation}
is obtained by propagating messages through the graph, starting from the terminal nodes and continuing until the messages $\mu_{c3}(x_3)$ and $\mu_{e3}(x_3)$ meet at edge $x_3$. These two messages are best understood as local evidence terms about $x_3$ accumulated along the respective message paths. Analogous to Bayes rule, the model's full belief about $x_3$ is obtained by multiplying the two incoming messages and, if needed, normalizing the result.

Summarizing, Bayesian inference in sparsely connected models, i.e., 
factorized models in which each factor depends only on a relatively small 
subset of variables, can be efficiently implemented through message passing 
on an FFG. The resulting message-passing procedure requires only local 
computations at the nodes.

\subsection{Constrained Variational Inference on Factor Graphs}\label{sec:cbfe}

The goal of this section is to show that the message-passing algorithm derived intuitively in Section~\ref{sec:forney-factor-graphs} can be recovered as a stationary solution of constrained VFE minimization on a factor graph. This result establishes message passing not as a computational heuristic but as principled variational inference.

Consider a generative model that factorizes as
\begin{equation}\label{eq:factorized-p}
p(x) = \prod_{a \in \mathcal{V}} f_a(x_a),
\end{equation}
with associated VFE functional
\begin{equation}\label{eq:VFE-for-factorized-p}
    F[q] =
    \int q(x)\log\frac{q(x)}{\prod_{a} f_a(x_a)} \,\mathrm{d}x\,,
\end{equation}   
where $x$ denotes a collection of variables and $x_a \subseteq x$ denotes the subset of variables associated with factor $f_a$. 

We now impose a specific factorization on $q(x)$, known as the \emph{Bethe constraint},\footnote{Each variable $x_i$ that appears in multiple factors would be overcounted by the product $\prod_a q_a(x_a)$. The inverse factors $q_i(x_i)^{-1}$ correct this overcounting.}
\begin{equation}\label{eq:bethe-constraint}
q(x) = \prod_a q_a(x_a)\prod_i q_i(x_i)^{-1},
\end{equation}
which introduces local beliefs that mirror the structure of the generative model: a belief $q_a(x_a)$ for each factor node and a belief $q_i(x_i)$ for each variable node \citep{yedidiaconstructing2005}. Substituting \eqref{eq:bethe-constraint} into \eqref{eq:VFE-for-factorized-p} leads to the \emph{Constrained Bethe
Free Energy} (CBFE) functional (see \citep[Eq.~7]{senozvariational2021a}):
\begin{equation}
F_{\mathrm{CBFE}}[q] =
\sum_a \int q_a(x_a)\log\frac{q_a(x_a)}{f_a(x_a)}\,\mathrm{d}x_a
+
\sum_i 
\int q_i(x_i)\log q_i(x_i)\,\mathrm{d}x_i \,,
\end{equation}
where the factorized beliefs are required to satisfy the following constraints:
\begin{subequations}\label{eq:bethe-marginalization-and-normalization}
\begin{align}
\int q_a(x_a)\,\mathrm{d}x_a &= 1 
\quad \forall a 
&&\text{(normalization)} \\
\int q_a(x_a)\,\mathrm{d}x_{a\setminus i} &= q_i(x_i)
\quad \forall (a,i) 
&&\text{(marginalization)} \, .
\end{align}
\end{subequations}

The Bethe constraint enables local message passing–based variational inference in models with factorized structure \eqref{eq:factorized-p}. The additional constraints \eqref{eq:bethe-marginalization-and-normalization} enforce consistency between neighboring factor and variable beliefs.

Introducing Lagrange multipliers for each consistency constraint and taking functional derivatives of the CBFE yields local stationary solutions of the form \citep[Theorem 1]{senozvariational2021a}
\begin{subequations}\label{eq:VFE-stationary}
\begin{align}
q_a(x_a) &\propto f_a(x_a) \prod_{i \in a}\mu_{ia}(x_i) \quad &&\text{(node beliefs)}\\
q_i(x_i) &\propto \prod_{a \in i} \mu_{ai}(x_i)\quad &&\text{(edge beliefs)}\,,
\end{align}
\end{subequations}
where the messages $\mu_{ai}(x_i)$ and $\mu_{ia}(x_i)$ are given by
\begin{subequations}\label{eq:VFE-messages}
\begin{align}
\mu_{ai}(x_i) &= \int f_a(x_a) \prod_{j\in a\setminus i} \mu_{ja}(x_j) \,\mathrm{d}x_{a\setminus i}
\quad&&\text{(node $\rightarrow$ edge)}\\
\mu_{ia}(x_i) &= \prod_{b\in i\setminus a} \mu_{bi}(x_i)
\quad&&\text{(edge $\rightarrow$ node)} \,.
\end{align}
\end{subequations}

Equations \eqref{eq:VFE-stationary} and \eqref{eq:VFE-messages} recover the \emph{sum-product algorithm} that we illustrated in Section~\ref{sec:forney-factor-graphs}. In this sense, message passing can be understood as arising from constrained variational inference rather than from algebraic manipulation of the distributive law. The CBFE perspective is more general, because it naturally accommodates additional constraints on the inference task. 

\citet{senozvariational2021a} show that nearly all known message passing variants (sum-product/belief-propagation, structured and mean-field variational message passing, data-constrained sum-product, Laplace propagation, expectation propagation) can be derived from first principles by varying local constraints on the variational posteriors $q_a(x_a)$ and $q_i(x_i)$. Two types of constraints are considered: (i)~factorization constraints (structured mean-field, naive mean-field), and (ii)~form constraints (data/Dirac-delta constraints, Laplace approximation, moment-matching for expectation propagation). CBFE minimization by local message passing thus provides a principled framework for trading off computational complexity and approximation accuracy in variational inference tasks. This flexibility is central for physical AI: the engineer (or the agent) can tune local approximations where time, energy, or memory are scarce, without abandoning the common VFE-minimizing architecture.

\subsection{Reactive Message Passing and RxInfer}\label{sec:reactive-message-passing}

\citet{bagaevreactive2023a} extend CBFE-based message passing with the concept of \emph{reactive programming}. In a Reactive Message Passing (RMP) framework, each node in the factor graph acts as an autonomous computational unit whose updates are scheduled locally in response to incoming changes. Message updates are considered only when incoming messages change, allowing inference to proceed through local event-driven updates. For example, when a new sensor value changes only one local likelihood factor, only the neighboring part of the graph needs to update immediately rather than the entire model. In this way, CBFE reduction is pursued through distributed local computations, although practical behavior still depends on the graph structure, update schedule, and approximation choices. Moreover, each node can suppress uninformative messages (e.g., those with near-uniform distributions) on-the-fly, thereby reducing computational load.

This reactive CBFE-minimizing message passing framework has been implemented in the open-source Julia toolbox \texttt{RxInfer}\footnote{\url{https://rxinfer.com}} \citep{bagaevrxinfer2023}. In \texttt{RxInfer}, practitioners specify a generative model together with a set of variational constraints, thereby defining a CBFE functional. When the conditions discussed in previous sections are satisfied, this functional corresponds to the VFE objective of an AIF agent. \texttt{RxInfer} can then automatically minimize the resulting free energy through a continual reactive message passing protocol. To give a concrete impression of the workflow, the following pseudocode sketches how a simple state-space model is specified and inferred in \texttt{RxInfer}:
\begin{verbatim}
  @model function state_space(y, T)
      x[0] ~ Normal(mean=0.0, var=10.0)       # prior on initial state
      for t in 1:T
          x[t] ~ Normal(mean=x[t-1], var=1.0)  # state transition
          y[t] ~ Normal(mean=x[t],   var=0.5)  # observation model
      end
  end
  result = infer(model=state_space(), data=(y=observations,))
\end{verbatim}
\noindent The \texttt{@model} macro builds the factor graph; \texttt{infer($\cdot$)} runs reactive message passing to return posterior beliefs over all latent states.

In summary, factor graphs provide a parallel, distributed architecture for minimizing variational free energy. Each node performs only local computations, and collective message passing across the graph solves the global inference problem. The CBFE formulation makes this framework flexible, because constraints on the variational family can be specified locally at individual nodes and edges, enabling principled trade-offs between inference accuracy and computational cost. Reactivity adds robustness and autonomy: since each node responds independently to incoming messages, inference can proceed uninterrupted even when data arrive asynchronously, sensors fail, or computational resources fluctuate. The next section explains why this inference paradigm is particularly well suited to the resource constraints faced by physical agents.

\section{Physical AI Agents as Active Inference Agents}\label{sec:physical-ai-as-aif}

The preceding sections have developed two complementary arguments.
Section~\ref{sec:FEP-and-AIF-agents} showed that the FEP provides a normative, first-principles design framework for
embodied agents: under the assumptions of the FEP, systems that maintain their structural and functional
integrity over time can be described as if they minimize variational free
energy. AIF operationalizes this principle by unifying
perception, learning, planning, and control within a single computational
objective, VFE minimization, without requiring separate mechanisms for
each. Section~\ref{sec:factor-graphs} showed that VFE minimization can
be realized efficiently and in a distributed manner by reactive message
passing on a factor graph, where each node autonomously performs only local
computations and the network collectively solves the global inference
problem. Together, these results suggest that \emph{the AIF framework, realized
by reactive message passing on a factor graph, offers a principled
foundation for physical AI agent design}.

\subsection{Continual Reactive Message Passing for Robustness}\label{sec:rmp-physical}

To appreciate why this matters for physical AI, consider the fluctuating operating
conditions that real-world physical AI devices face as a matter of course,
not as exceptional corner cases:

\begin{itemize}

\item \textbf{Temporal (deadline) fluctuations.}
Inference must complete within the time available before a decision is
required. An autonomous vehicle estimating the trajectory of an
oncoming car may have only tens of milliseconds before a collision
becomes unavoidable. The inference algorithm cannot request more time;
it must commit to the best available estimate when the deadline
arrives.

\item \textbf{Data fluctuations.}
Sensor data arrives sequentially and asynchronously across multiple
modalities. A robot tracking $N$ nearby objects receives, on average,
only $1/N$ of its sensor bandwidth per object, and some objects may
temporarily leave the field of view entirely. The inference algorithm
must incorporate each observation as it arrives, without waiting for a
complete synchronized snapshot.

\item \textbf{Power fluctuations.}
The computational budget per inference step is finite and
time-varying. A drone tracking $K$ targets can afford roughly $1/K$
of its processing capacity per target, and that budget shrinks further
as battery charge decreases. The inference algorithm must degrade
gracefully, trading accuracy for speed, rather than failing
abruptly when resources are insufficient.

\item \textbf{Compositional fluctuations.} The composition of the environment changes continuously. An agent navigating traffic may need to track anywhere from $2$ to $20$ other road users as vehicles, cyclists, and pedestrians enter and leave the scene. Equipment failures, such as a sensor dropping offline, further alter the effective model structure. Consequently, the agent’s generative model must adapt online to reflect these changing environmental compositions.
  
\end{itemize}

These are not incidental engineering difficulties; they are defining
characteristics of embodied, real-time operation. A principled
robust architecture must handle all four issues simultaneously, without requiring
the designer to anticipate every combination of conditions in advance.

Continual reactive message passing on a factor graph is well matched to this
challenge. Because each node responds to arriving messages without
requiring a global schedule, inference is event-driven: updates
proceed as soon as new data arrive and pause when no new information
is available. Hard temporal deadlines can be handled by committing to the current
beliefs at the required moment, regardless of whether message passing
has converged. Asynchronous or missing observations are absorbed
locally without affecting the rest of the graph. Reduced computational
resources simply mean that fewer messages are exchanged per unit time,
and since each completed local update is designed to reduce the CBFE, the system trades accuracy for speed in a principled fashion.

The CBFE framework further allows the complexity of inference to be
tuned locally at each node by choosing appropriate variational
constraints (Section~\ref{sec:cbfe}). Nodes operating under tight
resource budgets can adopt cheaper mean-field approximations, while
nodes with ample resources can use more expressive structured VMP or
expectation propagation updates. This local adaptability requires no
global re-engineering of the inference algorithm.

Crucially, the required computational resources cannot be scheduled in
advance: plans must be continuously updated when the environment
deviates from expectations,\footnote{``Everybody has a plan until they
get hit.'' ---Mike Tyson} and such deviations are by definition
unanticipated.

For example, state updating may need to operate at
$1\,\mu\mathrm{W}$, $1.1\,\mu\mathrm{W}$, or $1.2\,\mu\mathrm{W}$, and
produce updates within $1\,\mathrm{ms}$, $1.1\,\mathrm{ms}$, and so on.
Precomputing separate filter variants for all such conditions is
infeasible. Within the RMP framework, adaptation to such variation can
be handled locally: because computation is distributed across autonomous
nodes, the quality and frequency of updates can adjust to the resources
actually available at run time, without switching to a different global
inference architecture.

AIF agents realized in this way therefore inherit properties that are
directly relevant for physical deployment:
\begin{itemize}
    \item \emph{Unified design:} perception, learning, planning, and
    action selection all reduce to VFE minimization in a single
    generative model, eliminating the need to integrate separately
    engineered subsystems.
    \item \emph{Anytime inference:} reactive message passing can be
    interrupted at any point and returns the best available beliefs,
    making hard real-time deadlines tractable without special-purpose
    scheduling.
    \item \emph{Fault tolerance:} local autonomy can allow node
    failures or missing sensors to remain localized, so that performance
    degrades more gracefully than in tightly centralized architectures.
    \item \emph{Resource adaptability:} the accuracy--cost trade-off
    is controlled locally via variational constraints, allowing the
    agent to operate across a wide range of computational budgets
    without architectural changes.
\end{itemize}

\subsection{Computational Homogeneity}\label{sec:computational-homogeneity}

A further architectural consequence deserves emphasis. The framework
developed in this paper admits \emph{nested} realizations of AIF agents
without introducing a new computational primitive at higher levels of
organization. No level of the hierarchy introduces a different
computational mechanism: the only operations that appear, at every
scale, are the message computations of \eqref{eq:VFE-messages}.

This \emph{computational homogeneity} has a suggestive implication for
hardware design. A processing element that implements
\eqref{eq:VFE-messages} could serve as a reusable building block for a wide range of AIF realizations. Tiling such
elements and connecting them according to the factor graph topology may be sufficient to realize an AIF agent of arbitrary complexity. No
separate control logic, scheduler, or global inference engine is
needed alongside the message-passing substrate. The required silicon
operations are the same whether the agent is a single sensor node or
a large multi-modal robotic system. To illustrate, a conventional RL-based robot typically combines a convolutional perception module, a tree-search or model-predictive planner, a PID controller, and a policy-gradient learning algorithm, each requiring different computational primitives, software stacks, and integration interfaces. In the AIF framework, all of these functions reduce to message computations of the same form \eqref{eq:VFE-messages}.

Reactive message passing on a factor graph is therefore not merely a
convenient implementation strategy. It is the computational
architecture that matches the structure of VFE minimization to the
structure of real-world constraints, and it provides a homogeneous
substrate that scales from individual processing elements to full
agent hierarchies without changing the underlying computational
primitive.

\subsection{Illustration: An Active Inference Robot Football Team}\label{sec:aif-robot-football}

We now sketch how the framework developed in this paper applies to a team of robot football players. The goal is not an engineering specification but a concrete demonstration that VFE minimization, EFE-based planning, reactive message passing, and nested AIF agents compose naturally into a coherent physical AI architecture.

\subsubsection{The individual player as an AIF agent}\label{sec:player-aif}

Each robot player is an AIF agent with partition $(y_i, x_i, u_i, s_i)$. Following Section~\ref{sec:nested-aif}, we decompose each player’s states into outward-facing components (interfacing with the pitch, ball, and opponents) and inward-facing components (mediating inter-player coordination):
\begin{itemize}
    \item \emph{Sensory states:} $y_i^{\mathrm{out}}$ comprises camera, lidar, and proprioceptive signals that observe the environment; $y_i^{\mathrm{in}}$ comprises observed positions, movements, and gestures from teammates.
    \item \emph{External states:} $x_i^{\mathrm{out}}$ comprises ball position and velocity, opponent positions, and pitch boundaries; $x_i^{\mathrm{in}}$ comprises shared inter-player states such as coordinated passing lanes and relative teammate positioning.
    \item \emph{Control states:} $u_i^{\mathrm{out}}$ comprises locomotion and kicking actions that affect the ball and environment; $u_i^{\mathrm{in}}$ comprises gestures broadcast to teammates (e.g., signaling pass readiness or calling for the ball).
    \item \emph{Internal states:} $s_i$ encodes posterior beliefs about all external states (ball trajectory, opponent intentions, teammate plans).
\end{itemize}
Perception, learning, planning, and control for player $i$ all reduce to VFE minimization (Section~\ref{sec:state-inference}) under a single probabilistic generative model $p(y_i,x_i,u_i,s_i)$ that expresses assumed relationships between the above variables.

At the macro-level of a football player, policy selection is governed by minimization of the EFE $G(u_i)$ (Section~\ref{sec:control}), which gives each player two complementary drives. The \emph{goal-driven} term favors policies that move toward preferred states (scoring position, ball recovery, defensive cover). The \emph{epistemic} term favors policies that reduce uncertainty: a player with an occluded view of the ball is driven to reposition for better visibility before committing to a goal-directed action. Exploitation and exploration are thus not separately engineered; they emerge from a single variational objective.

\begin{figure}[t]
\centering
\resizebox{\columnwidth}{!}{%
\begin{tikzpicture}[>=Stealth, every node/.style={font=\small}]

\fill[red!5, rounded corners=5pt] (-9.0, -6.0) rectangle (9.0, 3.8);
\draw[red!30, rounded corners=5pt, thick] (-9.0, -6.0) rectangle (9.0, 3.8);
\node[font=\normalsize\bfseries, red!60!black] at (0, 4.2)
    {Environment (pitch, ball, opponents)};

\node[circle, draw=red!60!black, fill=red!10, very thick,
      minimum size=0.9cm, font=\footnotesize] (x1out) at (-8.2, -1.6) {$x_1^{\mathrm{out}}$};
\node[circle, draw=red!60!black, fill=red!10, very thick,
      minimum size=0.9cm, font=\footnotesize] (x2out) at (0, 3.0) {$x_2^{\mathrm{out}}$};
\node[circle, draw=red!60!black, fill=red!10, very thick,
      minimum size=0.9cm, font=\footnotesize] (x3out) at (8.2, -1.6) {$x_3^{\mathrm{out}}$};

\fill[teal!5, rounded corners=5pt] (-6.8, -4.8) rectangle (-3.2, 1.6);
\draw[teal!30, rounded corners=5pt, thick] (-6.8, -4.8) rectangle (-3.2, 1.6);
\node[font=\small\bfseries, teal!60!black] at (-5.0, -5.2) {Player $P_1$};

\draw[dashed, gray!55, line width=0.9pt] (-5.0, -1.6) ellipse (1.1cm and 2.5cm);

\node[circle, draw=gray!60, fill=gray!12, very thick,
      minimum size=0.8cm, font=\footnotesize] (y1) at (-5.0, 0.7) {$y_1$};
\node[circle, draw=teal!60!black, fill=teal!10, very thick,
      minimum size=0.8cm, font=\footnotesize] (s1) at (-5.0, -1.6) {$s_1$};
\node[circle, draw=gray!60, fill=gray!12, very thick,
      minimum size=0.8cm, font=\footnotesize] (u1) at (-5.0, -3.8) {$u_1$};

\draw[->, thick, teal!65!black] (y1) -- (s1);
\draw[->, thick, teal!65!black] (s1) -- (u1);

\node[circle, draw=orange!60, fill=orange!10, very thick,
      minimum size=0.9cm, font=\footnotesize] (x12) at (-2.5, -1.6) {$x_{12}^{\mathrm{in}}$};
\node[font=\footnotesize\bfseries, orange!60!black] at (-2.5, -2.6) {shared};

\fill[teal!5, rounded corners=5pt] (-1.3, -4.8) rectangle (1.3, 1.6);
\draw[teal!30, rounded corners=5pt, thick] (-1.3, -4.8) rectangle (1.3, 1.6);
\node[font=\small\bfseries, teal!60!black] at (0, -5.2) {Player $P_2$};

\draw[dashed, gray!55, line width=0.9pt] (0, -1.6) ellipse (1.1cm and 2.5cm);

\node[circle, draw=gray!60, fill=gray!12, very thick,
      minimum size=0.8cm, font=\footnotesize] (y2) at (0, 0.7) {$y_2$};
\node[circle, draw=teal!60!black, fill=teal!10, very thick,
      minimum size=0.8cm, font=\footnotesize] (s2) at (0, -1.6) {$s_2$};
\node[circle, draw=gray!60, fill=gray!12, very thick,
      minimum size=0.8cm, font=\footnotesize] (u2) at (0, -3.8) {$u_2$};

\draw[->, thick, teal!65!black] (y2) -- (s2);
\draw[->, thick, teal!65!black] (s2) -- (u2);

\node[circle, draw=orange!60, fill=orange!10, very thick,
      minimum size=0.9cm, font=\footnotesize] (x23) at (2.5, -1.6) {$x_{23}^{\mathrm{in}}$};
\node[font=\footnotesize\bfseries, orange!60!black] at (2.5, -2.6) {shared};

\fill[teal!5, rounded corners=5pt] (3.2, -4.8) rectangle (6.8, 1.6);
\draw[teal!30, rounded corners=5pt, thick] (3.2, -4.8) rectangle (6.8, 1.6);
\node[font=\small\bfseries, teal!60!black] at (5.0, -5.2) {Player $P_3$};

\draw[dashed, gray!55, line width=0.9pt] (5.0, -1.6) ellipse (1.1cm and 2.5cm);

\node[circle, draw=gray!60, fill=gray!12, very thick,
      minimum size=0.8cm, font=\footnotesize] (y3) at (5.0, 0.7) {$y_3$};
\node[circle, draw=teal!60!black, fill=teal!10, very thick,
      minimum size=0.8cm, font=\footnotesize] (s3) at (5.0, -1.6) {$s_3$};
\node[circle, draw=gray!60, fill=gray!12, very thick,
      minimum size=0.8cm, font=\footnotesize] (u3) at (5.0, -3.8) {$u_3$};

\draw[->, thick, teal!65!black] (y3) -- (s3);
\draw[->, thick, teal!65!black] (s3) -- (u3);

\draw[dashed, black!50, line width=1.2pt, rounded corners=8pt]
    (-7.1, -5.5) rectangle (7.1, 1.9);
\node[font=\footnotesize, black!60, anchor=north west] at (-7.0, 1.85)
    {Team agent states};
\node[font=\footnotesize, red!60!black, anchor=west] at (0.6, 3.0)
    {$x_{\mathrm{team}} = \{x_i^{\mathrm{out}}\}$};

\draw[->, thick, red!55!black] (x1out) to[out=90, in=180] (y1);
\draw[->, thick, red!55!black] (u1) to[out=180, in=270] (x1out);

\draw[->, thick, red!55!black] (x2out) to[out=270, in=90] (y2);
\draw[->, thick, red!55!black] (u2) to[out=90, in=180] (x2out);

\draw[->, thick, red!55!black] (x3out) to[out=90, in=0] (y3);
\draw[->, thick, red!55!black] (u3) to[out=0, in=270] (x3out);


\draw[->, thick, orange!85] (u1) to[out=0, in=240] (x12);
\draw[->, thick, orange!85] (x12) to[out=60, in=180] (y2);
\draw[->, thick, orange!85] (u2) to[out=180, in=300] (x12);
\draw[->, thick, orange!85] (x12) to[out=120, in=0] (y1);

\draw[->, thick, orange!85] (u2) to[out=0, in=240] (x23);
\draw[->, thick, orange!85] (x23) to[out=60, in=180] (y3);
\draw[->, thick, orange!85] (u3) to[out=180, in=300] (x23);
\draw[->, thick, orange!85] (x23) to[out=120, in=0] (y2);

\end{tikzpicture}%
}
\caption{A robot football team as coupled AIF agents. Three players $P_1$, $P_2$, $P_3$, each an AIF agent with sensory states $y_i$, internal states $s_i$, and control states $u_i$. Each player connects to the external environment (pitch, ball, opponents) through outward-facing states $x_i^{\mathrm{out}}$ (red arrows). Adjacent players are coupled through shared inter-agent spaces $x_{12}^{\mathrm{in}}$ and $x_{23}^{\mathrm{in}}$ (orange). The orange arrows represent the inward-facing components of $y_i$ and $u_i$: each player's inward actions $u_i^{\mathrm{in}}$ (e.g., broadcast signals) enter the shared space, while inward observations $y_i^{\mathrm{in}}$ (e.g., received messages) flow back from it. Players $P_1$ and $P_3$ interact only indirectly, through $P_2$.}
\label{fig:football-team}
\end{figure}
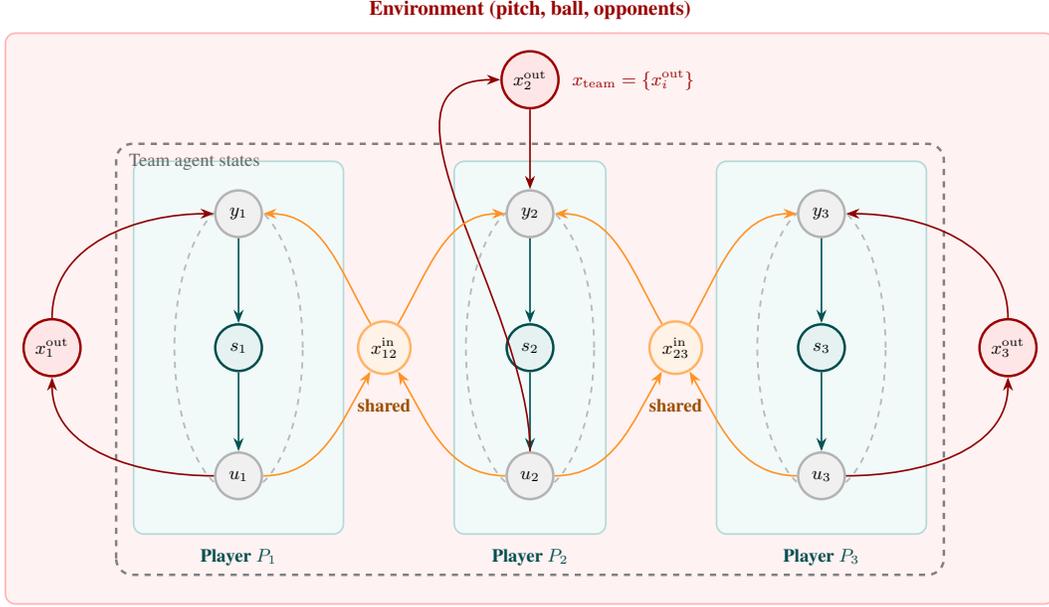

\subsubsection{The team as coupled AIF agents}\label{sec:team-nesting}

Consider three players $P_1$, $P_2$, $P_3$ coupled in a chain: $P_1$ and $P_2$ share a coupling link, as do $P_2$ and $P_3$, while $P_1$ and $P_3$ interact only indirectly through $P_2$ (Fig.~\ref{fig:football-team}). Inter-player coupling follows the E1 chain (Section~\ref{sec:nested-aif}),
\[
s_i \rightarrow u_i^{\mathrm{in}} \rightarrow x_i^{\mathrm{in}} \leftrightarrow x_j^{\mathrm{in}} \rightarrow y_j^{\mathrm{in}} \rightarrow s_j\,,
\qquad \text{for neighboring } i, j\,.
\]
For example, when $P_1$ intends to pass to $P_2$: the belief $s_1$ (pass intent) drives $u_1^{\mathrm{in}}$ (broadcast pass signal), which enters the shared inter-agent space $x_{12}^{\mathrm{in}}$, is received by $P_2$ as $y_2^{\mathrm{in}}$ (incoming pass message), and updates $P_2$’s belief $s_2$ to prepare for receiving the ball.

Each player minimizes its own VFE under the coupling induced by the shared inter-agent spaces. Coordinated behavior (spacing, passing sequences, role differentiation) emerges without any central controller. Each player’s EFE-based policy selection automatically accounts for the expected behavior of teammates, because the shared inter-agent states propagate belief updates across the chain.

This construction should be read as a schematic application of the
nesting conditions E1--E3 from Section~\ref{sec:nested-aif}, rather
than as a formal derivation for robot football. When those conditions
are satisfied, the coupled team admits a collective Markov-blanketed
partition. In Fig.~\ref{fig:football-team}, the team is represented as
the agent inside the dashed box around the three players. The
collective sensory states $y_{\mathrm{team}} = \{y_i^{\mathrm{out}}\}$
aggregate all players’ outward-facing observations; the collective
control states $u_{\mathrm{team}} = \{u_i^{\mathrm{out}}\}$ aggregate
all players’ outward-facing actions; and the collective internal states
$s_{\mathrm{team}}$ absorb individual beliefs $s_i$ together with all
inward-facing states. In that case, the team can itself be modeled as a
higher-level AIF agent that minimizes a team-level VFE, with team-level
EFE balancing collective goals against uncertainty about the opponent’s
strategy.

\subsubsection{Reactive message passing under fluctuating resources}\label{sec:football-rmp}

Football makes the resource constraints of Section~\ref{sec:physical-ai-as-aif} concrete. A player must decide to pass or shoot within hundreds of milliseconds (temporal constraint); observations are partial and asynchronous across teammates (data constraint); computational and motor budgets fluctuate with battery state and opponent count (power constraint). Continual reactive message passing handles all three simultaneously: inference is event-driven and can be committed to at any deadline; missing observations affect only the local graph region and cause beliefs to decay toward the prior; reduced resources mean fewer message iterations per second, corresponding to a coarser variational approximation, so that performance degrades smoothly rather than collapsing. The same mechanism scales to the team level without introducing any new computational primitive, which is precisely the architectural homogeneity argued for throughout this paper.

\section{Discussion}\label{sec:discussion}

\subsection{Context}\label{sec:context}

This paper is intended as a complement to the vision paper by
\citet{fristondesigning2022}, which argues that active inference provides a
first-principles foundation for designing ecosystems of natural and
artificial intelligence. That paper develops a roadmap of increasing
sophistication, from current function-approximation AI (stage~S0) through
sentient (S1), sophisticated (S2), sympathetic (S3), and shared (S4)
intelligence, and identifies factor-graph message passing as the
computational architecture that enables agents to share generative models
and coordinate through exchanged sufficient statistics.

The present paper addresses a question that \citet{fristondesigning2022}
deliberately leave open: how, concretely, should an engineer build the
individual agents that populate such an ecosystem?

Several specific contributions of the present paper fill gaps identified
in \citet{fristondesigning2022}. First, we show that reactive message
passing on a Forney-style factor graph directly addresses the real-time,
data, and power constraints that any physical agent faces
(Section~\ref{sec:physical-ai-as-aif}), constraints that
\citet{fristondesigning2022} acknowledge as fundamental but do not treat
in engineering detail. Second, the computational homogeneity result of
Section~\ref{sec:physical-ai-as-aif}, that every node in the factor graph
executes the same VFE-minimizing message computations, provides the
architectural primitive from which Friston et al.'s multi-agent ecosystems
can be assembled without introducing new mechanisms at each scale. Third,
the Constrained Bethe Free Energy framework of Section~\ref{sec:cbfe}
gives the engineer a concrete knob, namely the choice of local variational
constraints, for trading accuracy against computational cost at each node,
an essential capability for edge devices operating under the energy budgets
that \citet{fristondesigning2022} highlight via Landauer's principle.

A complementary perspective is offered by \citet{dupouxwhy2026}, who
identify autonomous learning as the central unsolved problem for physical
AI and propose a three-component architecture: observation-based learning
(System~A), action-based learning (System~B), and a meta-controller
(System~M) that routes data and switches learning modes based on
internally generated signals such as prediction error, novelty, and
uncertainty. We share their diagnosis: current AI systems cannot learn
autonomously in the way that biological organisms do, and closing this
gap requires tightly coupling perception, action, and an intrinsic drive
to explore.

The architectures differ, however, in a foundational commitment. The
framework of \citet{dupouxwhy2026} is deliberately agnostic about the
mathematical substrate: the learning objectives are generic loss
functions and expected-return maximizers, and
uncertainty enters only as scalar meta-signals (prediction error,
ensemble variance) that heuristically modulate data routing and
exploration. Probability theory plays no explicit role.

We argue that this agnosticism comes at a cost. An agent
that represents uncertainty only as a scalar ``surprise'' or ``novelty''
signal knows \emph{that} it is uncertain, but not \emph{about what}. To
decide which action would most effectively resolve its uncertainty, the
agent needs a structured, probabilistic representation of its beliefs
about the external world, because only then can it evaluate, for each
candidate policy, how much uncertainty that policy is expected to
resolve. This is precisely what the Expected Free Energy
(Section~\ref{sec:control}) provides: its ambiguity term
$\mathbb{E}_{q(x|u)}[\mathbb{H}[q(y|x)]]$ evaluates the expected
informativeness of future observations under each policy $u$, and this
evaluation is only meaningful when the agent maintains an explicit
posterior $q(x|u)$ over external states. Without such a posterior, the
EFE cannot be computed, and the agent is left to explore by heuristic
rather than by principled information seeking. In short, probability
theory is not merely a mathematical convenience; it is the substrate
that turns passive surprise into directed curiosity.

\subsection{Two Routes to Active Inference}\label{sec:two-routes}

Active inference and the Free Energy Principle can be approached by two
complementary routes, corresponding closely to the \emph{high road} and
\emph{low road} described by \citet{parractive2022}.

\emph{The high road.}
The high road is the physics and neuroscience route associated with
Friston's original derivation of the FEP
\citep{fristonfree2022,fristonpath2023}. In this route, the core result
that autonomous-state dynamics can be expressed as variational free
energy minimization is derived from assumptions about the physical
dynamics of a self-organizing system. In the present paper, this route
is reviewed in Section~\ref{sec:FEP}, culminating in the autonomous-state
dynamics of \eqref{eq:auto-states-dynamics} and their variational
reinterpretation in \eqref{eq:fep-vfe-def}.

\emph{The low road.}
The low road starts from probabilistic and inferential principles rather
than from physics. An important recent contribution is the work of
\citet{beckdynamic2025}, who derive the core FEP result from Jaynes'
principle of maximum caliber together with the Markov blanket
assumption. In their formulation, one again arrives at the conclusion
that updating of autonomous states can be expressed as variational free
energy minimization, but without assuming the physical postulates used
in Friston's derivation. In this sense, Beck and Ramstead clarify
\emph{what} the FEP is from an information-theoretic viewpoint: a
principled inferential law for systems with Markov-blanketed dynamics,
rather than necessarily a consequence of a particular physical starting
point.

The present paper also belongs to this low road, but with a different
emphasis. Whereas \citet{beckdynamic2025} provide an alternative
derivation of the FEP, our focus is on \emph{why} engineers working on physical
AI should pay attention to the FEP/AIF framework, and \emph{how} synthetic AIF
agents can be realized in practice. Our development follows the pathway
\[
\text{PT} \;\rightarrow\; \text{BML} \;\rightarrow\; \text{VI}
\;\rightarrow\; \text{AIF} \;\rightarrow\; \text{FFG/RMP},
\]
moving in steps from basic assumptions about rational reasoning under uncertainty
to the implementation of physical AI agents under real-world operating
conditions. 

The two routes are complementary rather than competing. The high road
provides a physics-based justification for why self-organizing systems
should be expected to minimize free energy. The low road shows that the
same principle can be understood from probability theory and inference,
and it offers the more direct entry point for engineers concerned with
building synthetic agents.

\begin{table*}[t]
\centering
\scriptsize
\setlength{\tabcolsep}{3pt}
\caption{Conceptual comparison between classical modular architectures for physical AI and the active-inference-based architecture argued for in this paper. Classical methods work well in stable, narrowly defined settings; the claim here is that active inference is a better architectural fit for embodied agents operating under fluctuating real-world constraints.}
\label{tab:aif-vs-classical}
\renewcommand{\arraystretch}{1.2}
\begin{tabular}{p{2.0cm} p{3.0cm} p{3.0cm} p{3.2cm}}
\toprule
\textbf{Aspect} & \textbf{Classical approach} & \textbf{AIF approach} & \textbf{Why it matters} \\
\midrule
Objective &
Separate objectives for perception, control, and learning &
Single variational free energy objective spanning perception, control, and learning &
Reduces fragmentation and ad hoc interfaces \\

Architecture &
Specialized modules with hand-designed couplings &
One generative model with local message passing updates &
Supports end-to-end coherence under uncertainty \\

Uncertainty &
Often split across estimator, planner, and controller &
Handled directly through probabilistic inference and (expected) free energy &
Keeps epistemic and goal-directed behavior in one formalism \\

Learning &
Commonly added as a separate subsystem or offline stage &
Can be formulated as active learning of parameters and, in principle, structure &
Aligns adaptation with goals and uncertainty \\

Exploration &
Typically heuristic or bonus-based &
Driven by risk, ambiguity, and novelty in expected free energy &
Avoids bolting exploration onto a separate control stack \\

Reward / preferences &
Externally specified reward function $R(x,u)$ &
Preference model $\hat{p}(x)$ within the generative model &
Avoids a separate reward-design layer \\

Asynchronous data &
Often handled through buffering or synchronization layers &
Updates are triggered locally when relevant messages change &
Matches irregular multimodal sensor streams \\

Resource constraints &
Modules are hardened separately; global scheduling is often needed &
Local variational constraints tune the accuracy--cost trade-off per node &
Supports graceful degradation under limited time, energy, or compute \\

Fault tolerance &
Failure in one module can cascade &
Local autonomy means failures degrade performance gracefully &
Improves robustness in embodied deployment \\

Real-time deadlines &
Requires per-module worst-case timing analysis &
Reactive message passing is interruptible and returns current beliefs &
Improves anytime operation under hard deadlines \\

Changing composition &
Adding or removing entities often requires model or software redesign &
Factor-graph structure can adapt locally as entities and relations change &
Better fit for open-world environments \\

Scalability / nesting &
New abstractions and interfaces are often introduced at each level &
The same message passing primitive can be reused across nested levels &
Supports computational homogeneity from components to teams \\

Hardware &
Typically optimized for centralized digital compute &
Naturally aligned with distributed, local, parallel computation &
Suggests a route toward neuromorphic-style implementations \\
\bottomrule
\end{tabular}
\end{table*}

\subsection{Active Inference vs.\ Reinforcement Learning}\label{sec:aif-vs-rl}

Table~\ref{tab:aif-vs-classical} summarizes the key architectural differences between classical modular AI systems and AIF realized by reactive message passing. Reinforcement learning (RL) and active inference both provide frameworks for designing agents that interact with environments to achieve goals. However, their differences extend beyond superficial distinctions. We highlight two critical issues: the reward function problem and computational homogeneity.

\emph{The reward function problem.}
In classical reward-centric RL pipelines, the agent's objective is typically specified through a reward function $R(x_t,u_t)$ or value criterion designed by the practitioner.\footnote{In RL, the agent's actual objective is not $R$ itself but the value function $V(x) = \mathbb{E}_u\left[\sum_{t=0}^{\infty} \gamma^t R(x_t, u_t) \right]$, which is the expected cumulative discounted reward under policy $u$.} This introduces two related difficulties. First, the handling of uncertainty and exploration is often introduced through additional modeling choices rather than being built into the core objective. This does not mean that RL cannot represent uncertainty or support exploration: Bayesian RL, POMDP-based control, and intrinsic-motivation methods all do so. The contrast is rather that these ingredients are often introduced as additional components or auxiliary objectives, whereas in active inference epistemic and goal-directed terms are combined within a single variational objective. Second, the reward functional still has to be designed by a human practitioner. Specifying a reward function that produces the desired behavior across the full range of operating conditions encountered in physical deployment is notoriously difficult and remains an open problem. Moreover, in standard RL the agent has no principled mechanism for resolving its own uncertainty about whether $R$ is correct, since $R$ is external to the inference process.

Active inference addresses these issues in a more unified way within a single probabilistic framework. Its cost function is the VFE $F[q]$, which does not encode any rewards explicitly but only quantifies the \emph{quality} of the agent's beliefs $q$ relative to a learned steady-state distribution $p$. This distribution is composed of multiple submodels, including a predictive model of the environment and a preference model that encodes rewarding future states, with parameters learned from experience. The cost function in AIF is therefore not hand-designed for each problem but is a fixed functional $F$ of the agent's beliefs $q(x)$ about the world, and the epistemic drive toward information-seeking behavior emerges automatically from the ambiguity term of the EFE (Section~\ref{sec:control}).

\emph{Computational homogeneity.}
A second, architectural difference concerns the uniformity of the computations an agent must perform. A typical RL system is composed of heterogeneous modules (a perception module, a world model, a policy network, a value estimator, and an exploration heuristic), each trained with a different objective and implemented with a different computational mechanism. Integration of these modules is a significant engineering challenge, and the interfaces between them are often the source of brittleness in physical deployment. Active inference, by contrast, enforces a full commitment to VFE minimization as the sole computational engine at every level of the agent hierarchy, with the engineering consequences discussed in Section~\ref{sec:computational-homogeneity}. This uniformity also carries a practical advantage. A physical agent must cope with temporal deadlines, asynchronous data arrivals, varying power budgets, and changing environmental composition (Section~\ref{sec:rmp-physical}), and in a modular RL system each component must be separately hardened against these fluctuations. When the computational engine is fixed as continual reactive message passing, many agent functions can be treated within the same inferential architecture, reducing the amount of separate engineering required to cope with these fluctuations.

\subsection{Active Inference, Active Learning, and Active Selection}\label{sec:active-selection}

So far, our discussion of synthetic AIF agent design has assumed that the
practitioner specifies the generative model, composed of a predictive model
and a preference model (see
\eqref{eq:generative-model-for-planning} where the epistemic priors are given by
\eqref{eq:epistemic-priors}).

These models may themselves contain unknown parameters. For example, a
predictive model $p'(y,x,u,\theta)$ may hold a set of parameters $\theta$ that needs to be fine-tuned over time. Within the FEP framework, VFE minimization
can be extended naturally to \emph{active learning} of such parameters
\cite{fristonfree2019}. In particular, \cite[Theorem~1]{vriesexpected2025}
show that augmenting the generative model~\eqref{eq:generative-model-for-planning} with an additional epistemic factor
\begin{equation}
    \tilde{p}(y,x) \propto \exp\!\Big(\mathrm{KL}\big[q(\theta | y,x)\,\|\,q(\theta | x)\big]\Big)
\end{equation}
induces an additional \emph{novelty} term
\begin{equation}
    \mathbb{E}_{q(x | u)}\big[I[\theta,y | x]\big]
\end{equation}
in the EFE, where $I[\theta,y | x]$ denotes the mutual
information between parameters $\theta$ and observations $y$, conditional on
external states $x$. This novelty term favors policies that are expected to
generate observations that reduce uncertainty about $\theta$, and therefore
support learning of the environment's dynamics.

Two points are worth emphasizing. First, active learning does not require an
ad hoc addition to the FEP framework; it emerges from the same variational
machinery as perception and control. Second, the learning is genuinely
\emph{active}: uncertainty reduction is balanced against the other components
of EFE, in particular risk and ambiguity. Because risk
captures the goal-directed aspect of behavior, the resulting learning pressure
is not generic curiosity in the abstract, but is shaped by the agent's
preferred outcomes. In that sense, active inference does not merely favor
learning per se; it favors learning that is useful for adaptive,
goal-directed behavior, and may therefore support more parsimonious models
than undirected exploration.

In biological systems, one may take this idea one step further and ask how the
structure of the generative model itself is acquired. On evolutionary time
scales, one may view this as a form of structure
learning under the FEP, with natural selection shaping the model class
available to an organism. In an engineering context, the
corresponding ideal would be that only high-level design constraints need to
be specified in a preference model, such as the desired future states associated with a cleaning
robot's task, while lower-level model structure is learned autonomously.

This motivates current work on \emph{active selection} of generative model
structures \cite{fristonsupervised2024,fristonactive2025a}. Here again,
the goal is to remain within the same VFE-minimization framework, but now to
use EFE to guide model selection and structure learning. Concretely, this
means choosing among candidate model structures, for example deciding
whether the dynamics of a ball in flight are better captured by a linear or a
nonlinear state-space model, or whether a latent variable representing wind
should be included, and selectively acquiring the data that is most useful for
disambiguating among these candidates. In this setting, active
selection is not just active data gathering in general, but data gathering in
the service of resolving uncertainty over model structure while remaining
sensitive to the agent's other objectives in the EFE, including risk, ambiguity, and
novelty. How such active selection strategies can be integrated into a
factor-graph message passing framework remains, to our knowledge, an open
question.

\subsection{Limitations}\label{sec:limitations}

The theoretical case for active inference as a foundation for physical AI is, we believe, solid. The arguments presented in this paper rest on probability theory, variational inference, and factor graph message passing, each of which is a mature and well-understood discipline. In this respect, the foundations of AIF are no less rigorous than those of reinforcement learning or optimal control.

The engineering case, however, remains largely unvalidated. Most of the advantages claimed in this paper, including anytime inference, principled exploration, and graceful degradation under resource constraints, have been demonstrated in small-scale experiments, but have not yet been stress-tested in the kinds of large-scale, real-time physical deployments where these properties would matter most. Closing the gap between theoretical promise and engineering practice is the central challenge facing the AIF community.

A concrete symptom of this gap is the current state of tooling. Realizing AIF agents in practice requires software infrastructure for specifying generative models, performing reactive message passing, and managing the computational graph at runtime. \texttt{RxInfer} (Section~\ref{sec:reactive-message-passing}) \citep{bagaevrxinfer2023} is the most mature open-source platform for this purpose and represents a significant step forward, but it is not yet at the level of robustness, documentation, and community support that engineers expect from production-grade tools. The absence of well-maintained, professionally supported toolboxes is a practical barrier to adoption that the field has not yet overcome.

A related limitation is the scarcity of engineering talent with the combined background in probabilistic inference, factor graphs, and real-time embedded systems that AIF agent development requires. The field currently draws primarily from theoretical neuroscience, philosophy and mathematical physics, communities whose research priorities and engineering norms differ substantially from those of robotics, signal processing, and control.

\section{Conclusions}\label{sec:conclusions}

This paper has argued that active inference provides a principled architectural framework for physical AI agents. Starting from probability theory, Bayesian machine learning, and variational inference, we showed how active inference extends these ideas to embodied agents that must perceive, learn, plan, and act under uncertainty in real time. From this perspective, variational free energy provides a unified computational objective, replacing the fragmented collection of separately engineered objectives that characterize many contemporary physical AI systems.

The second part of the argument concerned realization. We argued that reactive message passing on a factor graph provides a distributed computational architecture that is well matched to the constraints of physical deployment. Because computation is local, event-driven, and interruptible, this architecture is naturally suited to hard temporal deadlines, asynchronous data arrival, fluctuating power budgets, and changing environmental composition. The same message-passing primitive can also be reused across nested levels of organization, yielding a computationally homogeneous architecture from internal components to multi-agent systems.

The contribution of this paper is therefore not a benchmark study or a claim that large-scale engineering validation has already been achieved. Rather, it is to make the theoretical and architectural case for active inference accessible to an engineering audience, and to argue that this framework deserves serious consideration as a foundation for physical AI. If the persistent gap between current embodied AI systems and biological agents is to be narrowed, we believe that progress will require not only better implementations, but also better architectural principles. Active inference, we have argued, is a strong candidate for such a principle.

\section{Acknowledgments}

The author thanks the current and former members of BIASlab at Eindhoven University of Technology for years of collaborative work, the team at Lazy Dynamics BV for their pioneering efforts to bring Bayesian technology to real-world applications, and GN Advanced Science for their sustained commitment to fundamental research. This work was supported by GN Advanced Science, Eindhoven University of Technology, the Netherlands Enterprise Agency (RVO, programs TKI2112P09, NGFAI2507, NGFAI2502), and the Dutch Research Council (NWO, program KICH3.LTP.20.006).

\bibliographystyle{plainnat}
\bibliography{references-auto}

\appendix

\section{Derivation of Evidence and Posterior for the Coin Toss Model\texorpdfstring{~\eqref{eq:coin-toss-evidence-1}}{ (Eq. \ref{eq:coin-toss-evidence-1 })}}\label{app:coin-toss-proof}

We derive the evidence $p(D|m)$ and posterior $p(\mu|D,m)$ for the coin toss model of Example~\ref{ex:coin-toss}. The likelihood and prior are
\begin{align*}
    p(D|\mu) &= \mu^n(1-\mu)^{N-n} \,, \\
    p(\mu) &= \mathrm{Beta}(\mu|\alpha,\beta) = \frac{1}{B(\alpha,\beta)}\,\mu^{\alpha-1}(1-\mu)^{\beta-1}\,,
\end{align*}
where $B(\alpha,\beta) \triangleq \Gamma(\alpha)\Gamma(\beta)/\Gamma(\alpha+\beta)$. Multiplying likelihood and prior and combining exponents gives
\begin{align}
    p(D|\mu)\cdot p(\mu)
    &= \frac{1}{B(\alpha,\beta)}\,\mu^{n+\alpha-1}(1-\mu)^{N-n+\beta-1} \notag \\
    &= \underbrace{\frac{B(n+\alpha,\,N-n+\beta)}{B(\alpha,\beta)}}_{\text{evidence }p(D|m)}
       \cdot\underbrace{\frac{1}{B(n+\alpha,\,N-n+\beta)}\,\mu^{n+\alpha-1}(1-\mu)^{N-n+\beta-1}}_{\text{posterior }p(\mu|D,m)\,=\,\mathrm{Beta}(\mu\,|\,n+\alpha,\,N-n+\beta)}\,. \label{eq:coin-toss-derivation}
\end{align}
In the last step, the factor $B(n+\alpha, N-n+\beta)$ is inserted to normalize the posterior and compensated for in the evidence. The evidence therefore evaluates to the scalar
\begin{equation}\label{eq:coin-toss-evidence}
    p(D|m) = \frac{B(n+\alpha,\,N-n+\beta)}{B(\alpha,\beta)}\,,
\end{equation}
and the posterior is again Beta-distributed with updated shape parameters $n+\alpha$ and $N-n+\beta$, confirming that the Beta prior is conjugate to the Binomial likelihood.

\section{Proof of the Complexity-Accuracy Decomposition\texorpdfstring{~\eqref{eq:CA-decomp}}{ (Eq. \ref{eq:CA-decomp})}}\label{app:complexity-accuracy}
\begin{subequations}\label{eq:proof-surprise}
    \begin{align}
    \underbrace{ -\log p(D|m)}_{\text{surprisal}} &=  -\log p(D|m) \underbrace{\int p(\theta|D,m)\mathrm{d}\theta}_{=1}  \\
     &= \int p(\theta|D,m)\log \frac{1}{p(D|m)}\mathrm{d}\theta \\
     &= \int p(\theta|D,m)\log \underbrace{\frac{p(\theta|D,m)}{p(D|\theta,m) p(\theta|m)}}_{\text{by Bayes rule}}\mathrm{d}\theta \\
     &= \underbrace{\int p(\theta|D,m)\log \frac{p(\theta|D,m)}{ p(\theta|m)}\mathrm{d}\theta}_{\text{complexity}} - \underbrace{\int p(\theta|D,m)\log p(D|\theta,m)\mathrm{d}\theta}_{\text{accuracy ("data fit")}}
\end{align}
\end{subequations}

\section{Derivation of the EFE-based Planning as VFE Minimization\texorpdfstring{~\eqref{eq:VFE-for-control}}{ (Eq. \ref{eq:VFE-for-control})}}\label{app:vfe-control-derivation}

Substituting the steady-state density model $f(y,x,u) \propto p'(y,x,u)\hat{p}(x)\tilde{p}(x)\tilde{p}(u)$ into the VFE definition gives
\begin{subequations}\label{eq:VFE-as-EFE-proof}
\begin{align}
    F[q] &= \mathbb{E}_{q(y,x,u)}\Big[ \log \frac{q(y,x,u)}{f(y,x,u)}\Big] \\
    &= \mathbb{E}_{q(y,x,u)}\Big[ \log \frac{q(y,x,u)}{p'(y,x,u)\hat{p}(x)\tilde{p}(x)\tilde{p}(u)}\Big] \\
    &=  \mathbb{E}_{q(u)}\bigg[ \underbrace{\mathbb{E}_{q(y,x|u)}\Big[ \log \frac{q(x|u)}{\hat{p}(x)q(y|x)}\Big]}_{G(u)}\bigg] + \mathbb{E}_{q(y,x,u)} \Bigg[ \log \frac{q(y,x,u)}{p'(y,x,u)\cancel{\hat{p}(x)}\tilde{p}(x)\tilde{p}(u)} \cdot \underbrace{\frac{\cancel{\hat{p}(x)}q(y|x)}{q(x|u)}}_{1/G(u)}\Bigg] \\
    &= \mathbb{E}_{q(u)}\big[ G(u)\big] + \mathbb{E}_{q(y,x,u)}\Big[ \log \frac{q(y,x,u)}{p'(y,x,u)}\Big] + \underbrace{\mathbb{E}_{q(x,u)}\Big[\log \frac{1}{q(x|u)\tilde{p}(u)} \Big]}_{=0 \text{ if }\eqref{eq:u-epistemic-prior} \text{ holds}} + \underbrace{\mathbb{E}_{q(y,x)}\Big[\log \frac{q(y|x)}{\tilde{p}(x)} \Big]}_{=0 \text{ if }\eqref{eq:x-epistemic-prior} \text{ holds}} \label{eq:VFE-as-EFE-proof-epistemic-priors}\\
    &= \mathbb{E}_{q(u)}\Bigg[ G(u) + \underbrace{\mathbb{E}_{q(y,x|u)}\Big[ \log \frac{q(y,x|u)}{p'(y,x|u)} \Big]}_{C(u)}+ \log \frac{q(u)}{p'(u)}\Bigg] \\
    &= \mathbb{E}_{q(u)}\Bigg[ \log \frac{q(u)}{\exp\big(-G(u) - C(u) - P(u)\big)}\Bigg]
    \qquad \text{if }\eqref{eq:epistemic-priors} \text{ holds}\,,
\end{align}
\end{subequations}
where $G(u)$, $C(u)$, and $P(u)$ are defined in \eqref{eq:Gu-def}--\eqref{eq:Pu-def}.  The third step separates out $\hat{p}(x)q(y|x)/q(x|u)$ to isolate the EFE term $G(u)$.  The fourth step uses the epistemic priors \eqref{eq:epistemic-priors} to cancel the two underbraced terms.  The final expression is a KL divergence, yielding \eqref{eq:VFE-for-control}.

\end{document}